\documentclass[10pt,twocolumn,letterpaper]{article}

\usepackage{cvpr}      

\usepackage{graphicx}
\usepackage{amsmath}
\usepackage{amssymb}
\usepackage{booktabs}
\usepackage{lipsum}
\usepackage{booktabs}

\usepackage{cuted}
\usepackage[pagebackref,breaklinks,colorlinks,linkcolor={}]{hyperref}

\usepackage[capitalize]{cleveref}
\crefname{section}{Sec.}{Secs.}
\Crefname{section}{Section}{Sections}
\Crefname{table}{Table}{Tables}
\crefname{table}{Tab.}{Tabs.}

\def\cvprPaperID{6021} 
\def\confName{CVPR}
\def\confYear{2023}

\usepackage{xcolor}
\usepackage{xspace}
\usepackage[all]{nowidow}

\definecolor{turquoise}{cmyk}{0.65,0,0.1,0.3}
\definecolor{purple}{rgb}{0.65,0,0.65}
\definecolor{dark_green}{rgb}{0, 0.5, 0}
\definecolor{orange}{rgb}{0.8, 0.6, 0.2}
\definecolor{red}{rgb}{0.8, 0.2, 0.2}
\definecolor{darkred}{rgb}{0.3, 0.05, 0.025}
\definecolor{blueish}{rgb}{0.0, 0.3, .6}
\definecolor{light_gray}{rgb}{0.7, 0.7, .7}
\definecolor{pink}{rgb}{1, 0, 1}
\definecolor{greyblue}{rgb}{0.25, 0.25, 1}

\renewcommand{\paragraph}[1]{\vspace{.5em}\noindent\textbf{#1}.}

\usepackage{enumitem}
\setlist[itemize]{noitemsep,leftmargin=1.25em,topsep=.2em}
\setlist[enumerate]{noitemsep,leftmargin=1.25em,topsep=.2em}

\usepackage{lipsum}

\usepackage{currfile}

\newcommand{\eq}[1]{(\ref{eq:#1})}

\usepackage{environ}
\NewEnviron{scaletoline}{\resizebox{1.0\linewidth}{!}{\BODY}}

\newcommand{\image}{\mathbf{C}}
\newcommand{\iImage}{c}
\newcommand{\nImages}{C}
\newcommand{\point}{\mathbf{p}}
\newcommand{\nPoints}{P}
\newcommand{\iPoint}{p}
\newcommand{\iIter}{{(t)}}
\newcommand{\nIter}{T}
\newcommand{\iIterNext}{{(t+1)}}
\newcommand{\iIterPrev}{{(t-1)}}
\newcommand{\iIterZero}{{(0)}}
\newcommand{\rot}{\mathbf{R}}
\newcommand{\tra}{\mathbf{t}}
\newcommand{\expect}{\mathbb{E}}
\newcommand{\real}{\mathbb{R}}
\newcommand{\feature}{\mathbf{f}}
\newcommand{\dFeature}{F}
\newcommand{\featureglobal}{\mathbf{g}}
\newcommand{\encoder}{\mathcal{E}_{\text{init}}}
\newcommand{\posenc}{\boldsymbol{\gamma}}
\newcommand{\MLP}{\mathcal{N}_\text{init}}
\newcommand{\pars}{\boldsymbol{\theta}}

\begin{document}

\newcommand{\methodName}{Sparse-View Camera Pose Regression and Refinement\xspace}
\newcommand{\methodNameAbbrev}{SparsePose\xspace}

\title{SparsePose: Sparse-View Camera Pose Regression and Refinement}


\author{
Samarth Sinha$^{1}$\quad Jason Y. Zhang$^{2}$\quad \\ Andrea Tagliasacchi$^{1,3,4}$ Igor Gilitschenski$^1$\quad David B. Lindell$^{1,5}$
\\[1em]{\small $^1$University of Toronto\quad $^2$Carnegie Mellon University\quad $^3$Simon Fraser University\quad $^4$Google\quad $^5$Vector Institute}\vspace{-1em}}

\maketitle

\begin{abstract}

Camera pose estimation is a key step in standard 3D reconstruction pipelines that operate on a dense set of images of a single object or scene.
However, methods for pose estimation often fail when only a few images are available because they rely on the ability to robustly identify and match visual features between image pairs.  
While these methods can work robustly with dense camera views, capturing a large set of images can be time-consuming or impractical.
We propose \methodNameAbbrev for recovering accurate camera poses given a sparse set of wide-baseline images~(fewer than 10).
The method learns to regress initial camera poses and then iteratively refine them after training on a large-scale dataset of objects (Co3D: Common Objects in 3D).
\methodNameAbbrev significantly outperforms conventional and learning-based baselines in recovering accurate camera rotations and translations. 
We also demonstrate our pipeline for high-fidelity 3D reconstruction using only 5-9 images of an object.

\end{abstract}

\section{Introduction}
\label{sec:intro}

\begin{figure}
\centering
\includegraphics[width=\linewidth]{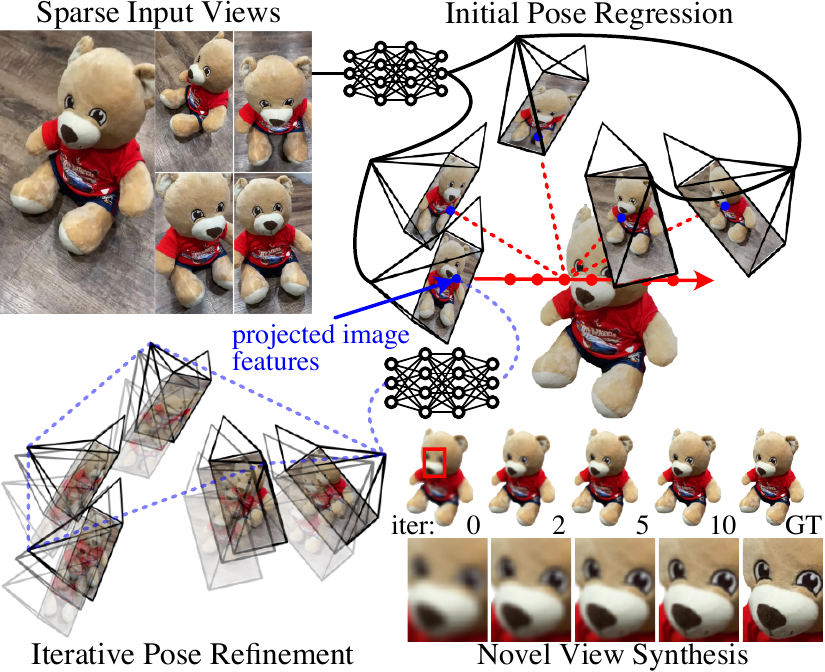}
\caption{\textbf{\methodNameAbbrev} -- Given sparse input views, our method predicts initial camera poses and then refines the poses based on learned image features aggregated using projective geometry. \methodNameAbbrev outperforms conventional methods for camera pose estimation based on feature matching within a single scene and enables high-fidelity novel view synthesis (shown for each iteration of pose refinement) from as few as five input views.
}
\label{fig:teaser}
\end{figure}
Computer vision has recently seen significant advances in photorealistic new-view synthesis of individual objects~\cite{nerf,co3d,anerf,codenerf} or entire scenes~\cite{mipnerf360,blocknerf,nerf++}.
Some of these multiview methods take tens to hundreds of images as input~\cite{nerf,lindell2022bacon,lindell2021autoint,mipnerf360}, while others estimate geometry and appearance from a few sparse camera views~\cite{co3d,Yu2020pixelnerf,niemeyer2022regnerf}.
To produce high-quality reconstructions, these methods currently require accurate estimates of the camera position and orientation for each captured image.







Recovering accurate camera poses, especially from a limited number of images is an important problem for practically deploying 3D reconstruction algorithms, since
it can be challenging and expensive to capture a dense set of images of a given object or scene.
While some recent methods for appearance and geometry reconstruction jointly tackle the problem of camera pose estimation, they typically require dense input imagery and approximate initialization~\cite{barf, nerfmm} or specialized capture setups such as imaging rigs~\cite{joo2015panoptic, humbi_dataset}. 
Most conventional pose estimation algorithms learn the 3D structure of the scene by matching image features between pairs of images \cite{sfm_revisited,sfm_survey}, but they typically fail when only a few wide-baseline images are available. The main reason for this is that features cannot be matched robustly resulting in failure of the entire reconstruction process.

In such settings, it may be outright impossible to find correspondences between image features. Thus, reliable camera pose estimation requires learning a \textit{prior} over the geometry of objects.  
Based on this insight, the recent RelPose method~\cite{relpose} proposed a probabilistic energy-based model that learns a prior over a large-scale object-centric dataset~\cite{co3d}.
RelPose is limited to predicting camera rotations (i.e., translations are not predicted). Moreover, it operates directly on image features without leveraging explicit 3D reasoning.



To alleviate these limitations, we propose \textit{\methodNameAbbrev}, a method that predicts camera rotation and translation parameters from a few sparse input views based on 3D consistency between projected image features (see Figure~\ref{fig:teaser}).
We train the model to learn a prior over the geometry of common objects~\cite{co3d}, such that after training, we can estimate the camera poses for sparse images and generalize to unseen object categories.
More specifically, our method performs a two-step coarse-to-fine image registration: 
(1) we predict \textit{coarse} approximate camera locations for each view of the scene, and (2) these initial camera poses are used in a pose refinement procedure that is simultaneously iterative and autoregressive,  which allows learning \textit{fine}-grained camera poses.
We evaluate the utility of the proposed method by demonstrating its impact on sparse-view 3D reconstruction.

Our method outperforms other methods for camera pose estimation in sparse view settings.  
This includes conventional image registration pipelines, such as COLMAP~\cite{sfm_revisited}, as well as recent learning-based methods, such as RelPose~\cite{relpose}.
Overall, \methodNameAbbrev enables real-life, sparse-view reconstruction with as few as five images of common household objects, and is able to predict
accurate camera poses, with only 3 source images of previously unseen objects.

In summary, we make the following contributions.
\begin{itemize}
    \item We propose \methodNameAbbrev, a method that predicts camera poses from a sparse set of input images. 
    \item We demonstrate that the method outperforms other techniques for camera pose estimation in sparse settings;
    \item We evaluate our approach on 3D reconstruction from sparse input images via an off-the-shelf method, where our camera estimation enables much higher-fidelity reconstructions than competing methods.
\end{itemize}


\section{Related Work}
\label{sec:rw}
Camera pose estimation from RGB images is a classical task in computer vision~\cite{moravec1980obstacle, ozyecsil2017survey}. 
It finds applications in structure-from-motion~(SfM~\cite{hartley2003multiple}), visual odometry~\cite{nister2004visual}, simultaneous localization and mapping (SLAM~\cite{slam}), rigid pose estimation~\cite{survey_rigid_pose}, and novel view synthesis with neural rendering~(NVS~\cite{nerf}).
In our discussion of related work, we focus on several related  types of pose inference as well as few-shot reconstruction.

\paragraph{Local pose estimation}
A variety of techniques estimate camera poses by extracting keypoints and \textit{matching} their local features across input images~\cite{superglue, siftflow, rootsift, loftr, pixelperfectsfm, requires_depth_multiview, jin2021image}. 
Features can be computed in a hand-crafted fashion~\cite{sift,bay2008speeded} or can be learnt end-to-end~\cite{lift, superpoint}.
Cameras are then refined via bundle adjustment, where camera poses and 3D keypoint locations are co-optimized so to minimize reprojection errors~\cite{bundle_adjustment, large_bundle_adjustment}.
Generally speaking, feature matching methods fail in \textit{few-shot} settings due to occlusion and limited overlap between images, where a sufficient number of common keypoints cannot be found.
These methods fail because they are \textit{local} -- and therefore do not learn \textit{priors} about the solution of the problem.

\paragraph{Global pose optimization}
Global pose optimization methods rely on differentiable rendering techniques to recover camera poses by minimizing a photometric reconstruction error~\cite{dove,lasr}.
Within the realm of neural radiance fields~\cite{nerf}, we find techniques for estimating pose between images~\cite{inerf} and between a pair of NeRFs~\cite{nerf2nerf}, as well as models that \textit{co-optimize} the scene's structure altogether with camera pose~\cite{barf,nerfmm,garf}, even including camera distortion~\cite{scnerf}.
SAMURAI \cite{boss2022samurai} is able to give very accurate camera poses by performing joint material decomposition and pose refinement,
but relies on coarse initial pose estimates, many images for training~($\approx$80), and trains from scratch for each new image sequence.
In contrast, our method is trained once and can then perform forward inference on unseen scenes in seconds.
Overall, such global pose optimization methods often fail in few-shot settings since they require sufficiently accurate pose initializations to converge. 
While local methods are often used for initialization, they too fail with sparse-view inputs, and so cannot reliably provide pose initializations in this regime.


\begin{figure*}
\centering
\includegraphics[width=\linewidth]{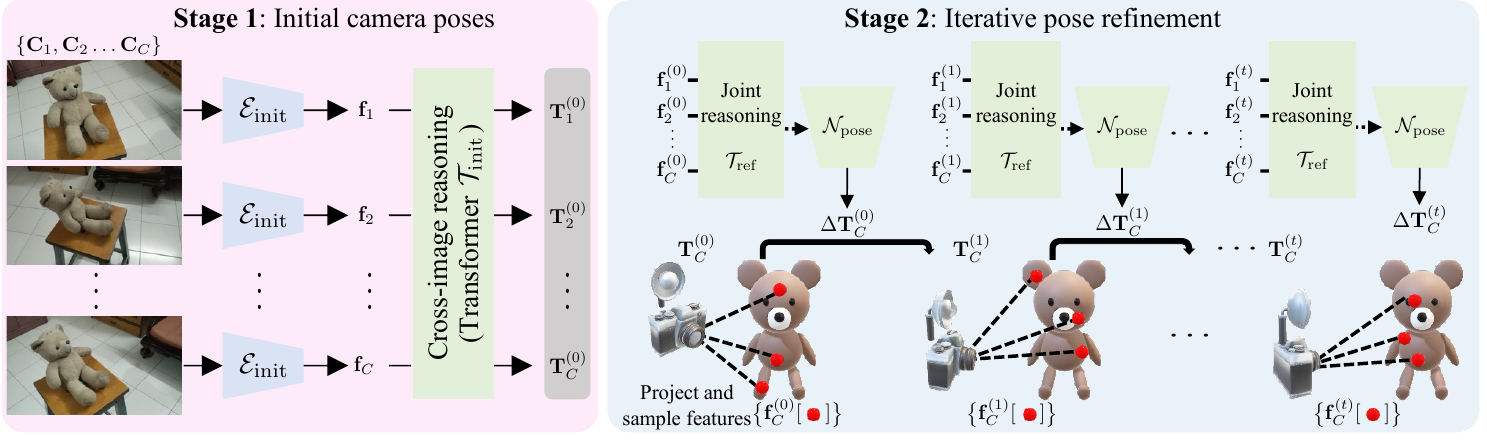}
\caption{\textbf{Method Overview.} We propose \methodName (\methodNameAbbrev), which takes as input few-views of an object from wide baselines, and predicts the camera poses.
\methodNameAbbrev is trained on a large scale dataset of ``common objects'' to learn a prior over the 3D geometry of the scene and the object.
Our method works by first predicting coarse initial camera poses by performing cross-image reasoning. 
The initial camera pose estimates are then iteratively refined in an auto-regressive manner, which learns to implicitly encode the 3D geometry of the scene based on sampled image features.
For notational convenience and simplicity, we use $\mathbf{T}$ to represent the rotations $\rot$ and translations $\tra$ in homogeneous coordinates (as used in the text).
}
\label{fig:method}
\end{figure*}

\begin{figure}
    \centering
    \includegraphics[width=\linewidth]{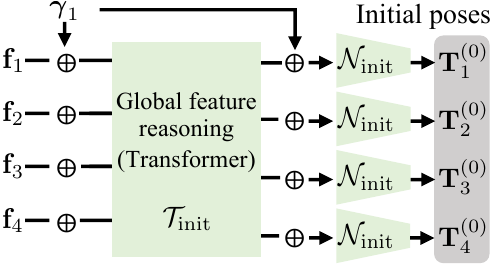}
    \caption{
    \textbf{Stage 1 architecture:} We initialize the camera poses by directly estimating the models using \textit{global reasoning}, and directly regressing the poses
    using pretrained features and joint-reasoning over the source images.
    We note that $\bigoplus$ denotes a skip connection (or addition) between the input and the output of the transformer $\mathcal{T}_{\text{init}}$,
    and for learnable positional encoding $\mathbf{\gamma}$. 
    For simplicity we use $\mathbf{T}$ to represent the rotations $\rot$ and translations $\tra$ in homogeneous coordinates (as used in the text).
    A detailed approach of stage 1 is in Section \ref{sec:init}.
    }
    \label{fig:sec1_method}
\end{figure}

\begin{figure}
    \centering
    \includegraphics[width=\linewidth]{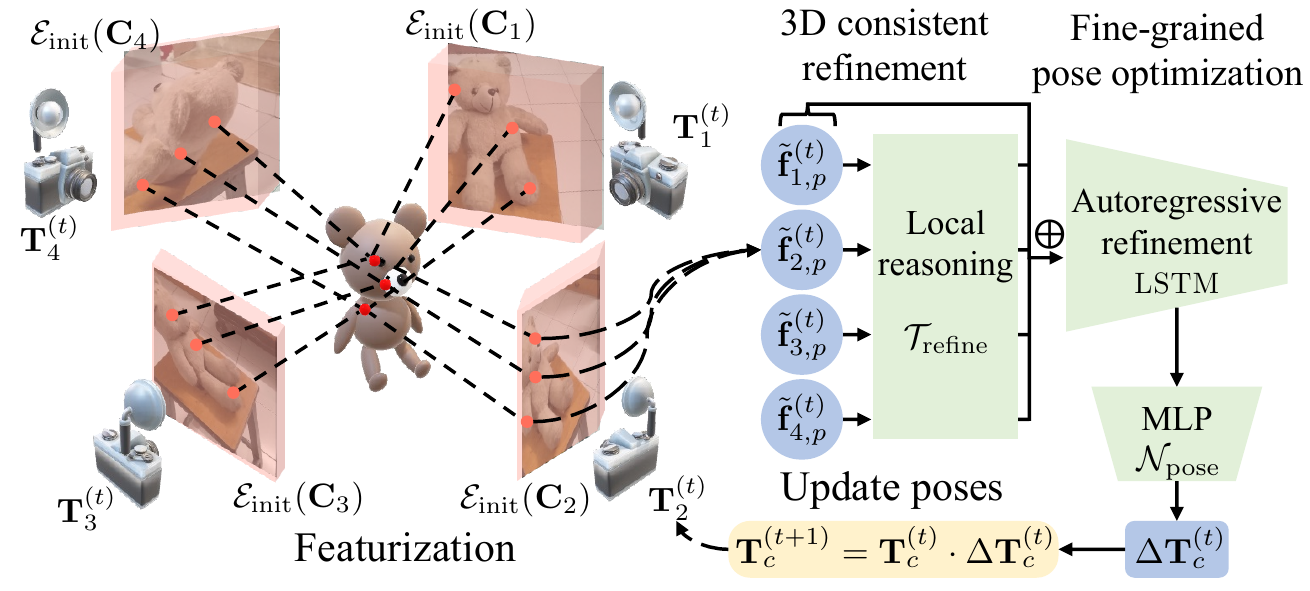}
    \caption{
    \textbf{Stage 2 architecture:} After obtaining the initial camera poses from Stage 1, we iteratively and autoregressively refine the camera poses using a local feature
    reasoning module, which learns the optimization dynamics of the camera poses. 
    Since the optimization is non-linear, the model iteratively updates the camera poses by resampling points and predicting pose offsets.
    We note that $\bigoplus$ denotes a skip connection between the input and the output of the transformer $\mathcal{T}_{\text{refine}}$, 
    and for simplicity we use $\mathbf{T}$ to represent the rotations $\rot$ and translations $\tra$ in homogeneous coordinates (as used in the text).
    A detailed approach of stage 2 is in Section \ref{sec:refine}.
    }
    \label{fig:sec2_method}
\end{figure}

\paragraph{Global pose regression}
Given a set of input images camera poses can also be directly regressed.
In visual odometry, we find techniques that use neural networks to auto-regressively estimate pose~\cite{deep_vo2, deep_vo3}, but these methods assume a small baseline between subsequent image pairs, rendering them unsuitable to the problem at hand.
\textit{Category-specific} priors can be learnt by robustly regressing pose w.r.t. a ``canonical'' 3D shape~\cite{ssd-6d, object_pose_1, object_pose_2}, or by relying on strong semantic priors such as human shape~\cite{metapose, vircoor} or (indoor) scene appearance~\cite{dirnet}.
Closest to our method, recent \textit{category-agnostic} techniques, such as RelPose~\cite{relpose}, are limited to estimating \textit{only rotations} by learning an energy-based probabilistic model
over $\mathrm{SO(3)}$.
However, RelPose only considers the global image features from the sparse views and does not perform local 3D consistency.
Even for predicting rotations, our method performs significantly better than RelPose since it takes into account both global and local features from images.



\paragraph{Direct few-shot reconstruction}
Rather than regressing pose and then performing reconstruction, it is also possible to reconstruct objects \textit{directly} from (one or more) images using data-driven category priors \cite{co3d}, or directly training on the scene.
Category-specific single-image 3D reconstruction estimate geometry and pose by matching pixels or 2D keypoints to a 3D template~\cite{c3dpo, ttp, csm, umr}, learning to synthesize class-specific 3D meshes~\cite{ucmr, umr}, or exploiting image symmetry~\cite{unsup3d, dove}.
Recent progress has also enabled few-shot novel view synthesis, where images of the scene from a novel viewpoint are generated \textit{conditioned} on only a small set of images~\cite{co3d,srt,threedim,enr, pixelnerf,niemeyer2022regnerf,jain2021putting,dsnerf, wce}.
Such methods are either trained to learn category-centric features \cite{co3d,wce,dove}, or are trained on a large-scale dataset to encode the 3D geometry of the scenes \cite{srt,threedim},
or propose regularization schemes for neural radiance based methods \cite{niemeyer2022regnerf,dsnerf, jain2021putting}.
However, 3D consistency in these models is learnt by \textit{augmentation} rather than by construction, resulting in lower visual quality compared to 
our approach.

\vfill
\section{Method}
\label{sec:method}


Estimating camera parameters typically involves predicting the intrinsics~(i.e. focal length, principal point, skew) and extrinsics~(i.e. rotation, translation) from a set of images.
In this paper, we only consider the task of estimating \textit{extrinsics}; we assume the intrinsics are known as they can be calibrated once for a camera a priori and are often provided by the camera manufacturer. 
More formally, our goal is to jointly predict the rotation $\rot_\iImage {\in} \mathrm{SO}(3)$ and translation~$\tra_\iImage {\in} \mathbb{R}^3$ for all input images~$\image_\iImage$.
Our proposed method for this task consists of two phases, which are illustrated in~\autoref{fig:method}:
\begin{itemize}
\item \textbf{\autoref{sec:init}} -- we first \textbf{initialize} the camera poses in a coarse prediction step which considers the \textit{global} image features in the scene.
\item \textbf{\autoref{sec:refine}} -- we \textbf{refine} the poses using an iterative procedure in which an autoregressive network predicts updates to align \textit{local} image features to match the 3D geometry of the scene.

\end{itemize}
%
The goal of the coarse pose estimation is to use global image features to provide an initial 3D coordinate frame and estimates of the relative camera rotations and translations which can then be iteratively refined.



\subsection{Initializing camera poses}
\label{sec:init}
%
Given a sparse set of $\nImages$ images $\{\image_1, \ldots, \image_\nImages\}$ of a single object, the first task is to predict initial camera poses and establish a coordinate frame.
The initialization:
\begin{itemize}
\item extracts low-resolution image features $\feature \in \real^\dFeature$;
\item combines image features into a global representation;
\item regresses rotation and translation for each camera.
\end{itemize}

We use a pre-trained, self-supervised encoder $\encoder$~\cite{dino} to extract features from each image.
Following VIT~\cite{vit}, we also add a learnable positional embedding $\posenc_\iImage \in \real^\dFeature$  to each feature:
%
\begin{equation}
\feature_\iImage = \overbrace{\encoder(\image_\iImage)}^\text{pre-trained features}
~~+
\overbrace{\rule{0pt}{1.6ex}\posenc_\iImage}^\text{learnable encoding}.
\end{equation}
These features are passed to a transformer $\mathcal{T}_{\text{init}}$~\cite{attention, vit} and a skip connection to aggregate global context and predict a new set of features:
\begin{equation}
\featureglobal_\iImage = \mathcal{T}_{\text{init}}(\{\feature_1, \ldots, \feature_\nImages\}; \pars) ~+~ \feature_\iImage.
\label{eq:initjoint}
\end{equation}
Finally, a shallow fully-connected network $\MLP$ (we use two hidden layers) predicts quaternions representing the initial camera rotations $\rot {\in} \mathrm{SO}(3)$, and translations $\tra{\in} \mathbb{R}^3$:
%
\begin{equation}
\rot^{(0)}_\iImage, \tra^{(0)}_\iImage = \MLP(\featureglobal_\iImage; \pars).
\end{equation}
The initial rotation and translation estimates are then refined using the iterative procedure described in \autoref{sec:refine}.

\subsection{Refining camera poses}
\label{sec:refine}
After obtaining the initial poses, $\rot^\iIterZero, \tra^\iIterZero$, we can leverage geometric reasoning to iteratively refine the pose estimates $\rot^\iIter, \tra^\iIter$, where $1 \leq t \leq \nIter$.
We achieve this by probing a collection of 3D points within the scene given the \textit{current} camera estimate (i.e.\ the points are re-sampled at each step). 
After projecting the points back into the images, features are fetched and aggregated into global feature vectors from which camera pose updates are computed.

\paragraph{Sampling}
We aim to uniformly sample within the volume where we expect the imaged object to be located.
Given the initially estimated camera poses, the center of the capture volume $\mathbf{c}$ is predicted considering principal rays~(i.e. rays passing through the principal point of each image).
We compute the point closest to the principal rays (in the least-squares sense) and  calculate the average camera radius as $r^\iIter = \expect_\iImage \| \mathbf{c} - \mathbf{c}_\iImage^\iIter \|_2$, where $\mathbf{c}_\iImage^\iIter$ is the camera center of image $\iImage$ at iteration $t$.
We then uniformly sample $\nPoints$ points within an Euclidean ball:
\begin{align}
\{\point^\iIter_\iPoint, \ldots, \point^\iIter_\nPoints\} \sim \mathcal{B}(\mathbf{c}^\iIter, r^\iIter/2).
\end{align}

To increase robustness of the optimization and to ensure the camera does not get stuck in a \textit{local minima}, we re-sample the 3D points after each camera pose update
to jitter the 3D points and image features, analogous to how PointNet jitters input pointcloud data \cite{qi2017pointnet}.

\paragraph{Featurization}
Let $\textbf{R}_\iImage^\iIter$ and $\textbf{T}_\iImage^\iIter$, denote the estimated rotation and translation at the $\iIter$-th iteration of the refinement procedure.
Given the set of 3D points and a known camera intrinsic matrix $\mathbf{K}$, we project them into the coordinate frame of each camera using 3D geometry:
\begin{align}
\point_{\iImage,\iPoint}^\iIter = \mathbf{\Pi}_{\iImage}^\iIter(\point_\iPoint^\iIter) = \mathbf{K}(\rot_\iImage^\iIter \point^\iIter_\iPoint + \tra_\iImage^\iIter).
\end{align}
We interpolate samples of the previously extracted image features $\feature$ at the projected 2D pixel coordinates for each source image~\cite{wce}, resulting in a set of feature embeddings for \textit{each} camera image and \textit{each} point at the \textit{current} refinement iteration.
We concatenate the positional encoding for the current predicted rotations and translations and the original 3D points to the embedding to generate a joint local feature vector:
%
\begin{align}
\feature_{\iImage,\iPoint}^\iIter &= \encoder(\image_\iImage)[\point_{\iImage,\iPoint}] \quad [\cdot] \equiv \text{ bilinear}
\\
\tilde{\feature}_{\iImage,\iPoint}^\iIter &= [
    \feature_{\iImage,\iPoint}^\iIter, \:\:
    \gamma(\point_\iPoint^\iIter), \:\:
    \gamma(\rot^\iIter_\iImage), \:\:
    \gamma(\tra^\iIter_\iImage) ] \in \real^{134},
\end{align}
where $\gamma(.)$ is a Fourier positional encoding~\cite{nerf}.
We then reduce the dimensionality of this large vector with a single linear layer $\mathcal{E}_\text{refine}$, and then concatenate along the \textit{samples} dimension:
\begin{equation}
\tilde{\feature}_\iImage^\iIter = \left[ 
    \mathcal{E}_\text{refine}(\tilde{\feature}_{\iImage,1}^\iIter; \pars),
    \ldots,
    \mathcal{E}_\text{refine}(\tilde{\feature}_{\iImage,\nPoints}^\iIter; \pars)
\right] \in \real^{\text{P} \cdot 32},
\end{equation}
where $\text{P}$ is the number of sampled points, which was chosen to be 1,000 resulting in a 32,000 dimensional local feature vector for each pose iteration step.
With this \emph{local} feature vector $\tilde{\feature}_\iImage^\iIter$ we summarize the appearance and geometry of the scene as sampled by the $\iImage$-th camera, allowing learned refinement of the camera poses in a 3D consistent manner to predict the camera pose updates.

\paragraph{Optimization}
Similar to \eq{initjoint}, we use a multi-headed self-attention module along with a skip connection to perform \textit{joint} reasoning over the source views:
%
\begin{equation}
    \tilde\featureglobal_\iImage = \mathcal{T}_\text{refine}(\tilde{\feature}_{0}^\iIter, \ldots,  \tilde{\feature}_\iImage^\iIter) + \tilde{\feature}_\iImage^\iIter,
\end{equation}
and regress pose updates using a long short-term memory network \textsc{lstm}~\cite{lstm} and a 2-layer MLP $\mathcal{N}_{\text{pose}}$:
\begin{gather}
\bar\featureglobal_\iImage^\iIter = \textsc{lstm}(\tilde\featureglobal_\iImage^\iIter; \{\tilde\featureglobal_\iImage^\iIterPrev, \dots, \tilde\featureglobal_\iImage^\iIterZero\})
\\
\begin{aligned}
\Delta \rot^\iIter_\iImage, \Delta \tra_\iImage^\iIter &= \mathcal{N}_{\text{pose}}\big(\bar\featureglobal_\iImage^\iIter\big) 
\\
\rot_\iImage^\iIterNext &= \rot_\iImage^\iIter \cdot \Delta \rot_\iImage^\iIter
\\
\tra_\iImage^\iIterNext &= \tra_\iImage^\iIter + \Delta \tra_\iImage^\iIter.
\end{aligned}
\end{gather}
Such auto-regressive models have been shown effective in implementing meta-optimization routines~\cite{learning2learn}, as they can learn priors on the dynamics of the optimization in few-shot settings \cite{lstm_fewshot}.
In practice, we perform 10 steps of the \textsc{lstm} pose refinement to allow for the camera poses to converge.
An additional ablation study for the number of steps is provided in the supplementary. 


To train the model we minimize the loss
\begin{align}
\mathcal{L}_{\text{pose}} &= \expect_\iImage \Sigma_{k\in \{0, K\}} \:\:
\mathcal{L}_{\text{pose}}^\rot + \mathcal{L}_{\text{pose}}^\tra,
\\
\mathcal{L}_{\text{pose}}^\rot &= d( \vert \vert \rot^\iIter_\iImage \rot^{\text{GT}}_\iImage - \mathbb{I} \vert \vert_{\mathcal{F}},
\\
\mathcal{L}_{\text{pose}}^\tra &= d(\vert \vert \tra_\iImage^\iIter - \tra_\iImage^{\text{GT}} \vert \vert_{2}^{2}),
\end{align}
where GT denotes ground truth data obtained by applying COLMAP~\cite{sfm_revisited} to \textit{densely} sampled video footage---note that we require dense frames at training time only, while our inference procedure uses sparse views.
To make the regression invariant to choices of coordinate frames, we align the ground truth rotation and translation such that the first source image is always at the unit camera location $\rot {=} \mathbb{I}$, $\tra {=} 0$.
The model then predicts \textit{relative} rotations and translations in this canonical coordinate space.
To stabilize training, we follow \cite{nsm, posenet} and take the loss over normalized quaternions.
For the penalty function $d(.)$ we use an adaptive and robust loss function \cite{robust_l2}.
The losses are applied only to the initial pose estimator and the output of the pose refinement module (i.e., the last iteration of the LSTM update).
Both prediction stages are trained jointly end-to-end.


%
%

\subsection{Implementation Details}
\paragraph{Training data}
We train the model on the CO3D dataset~\cite{co3d}, which contains 19,000 videos, with 1.5 M individual frames and camera poses across 50 different categories of common objects.
Training the model on this large dataset with diverse objects facilitates learning object-appearance priors, which ultimately enables pose prediction from sparse views.
We split the dataset into 30 train and 20 test categories, to verify the method's ability to adapt to novel classes.

\paragraph{Testing data}
To construct the test set, we use the 20 test categories, and sample 100 sequences for each number of source images $C \in [3,9]$.
To sample a test-set, we follow the \texttt{uniform}-variant of the evaluation protocol from RelPose~\cite{relpose}, and perform stratified sampling along the frame indices in the CO3D dataset to select for wide-baseline views~\cite{co3d}.
Furthermore, we use batch sampling from PyTorch3D \cite{pytorch3d} to shuffle batches with a random number of source images $C \in \lbrace 3,\ldots, 9\rbrace$,
so that the model learns how to aggregate features and jointly predicts camera poses with a different number of source images.
We note that the model architecture is designed to work with an arbitrary number of unposed source images. 
As mentioned previously, the camera intrinsic matrix $\mathbf{K}$ is assumed to be known for all training and testing sequences, which
is a reasonable assumption, given that the CO3D dataset was collected from smartphones and such parameters can be easily obtained from smartphone manufacturers.

\paragraph{Training details}
The model is trained on two A6000 48 GB GPUs for 3 days until convergence.
The model is jointly trained using an Adam optimizer \cite{adam} with initial learning rate of $10^{-4}$, which is decayed once by a factor of 10 after 
250 epochs of training.
The other Adam optimizer parameters are left to the default values from the PyTorch implementation \cite{paszke2019pytorch}.
A pre-trained DINO Vision Transformer is used to compute the pre-trained embeddings~$\mathcal{E}_{\text{init}}$ \cite{dino,vit}.
We use the frozen weights from the official release of the~\texttt{ViT-B/8} variant of the DINO ViT model.





\begin{figure}[t!]
    \centering
    \includegraphics[width=\linewidth]{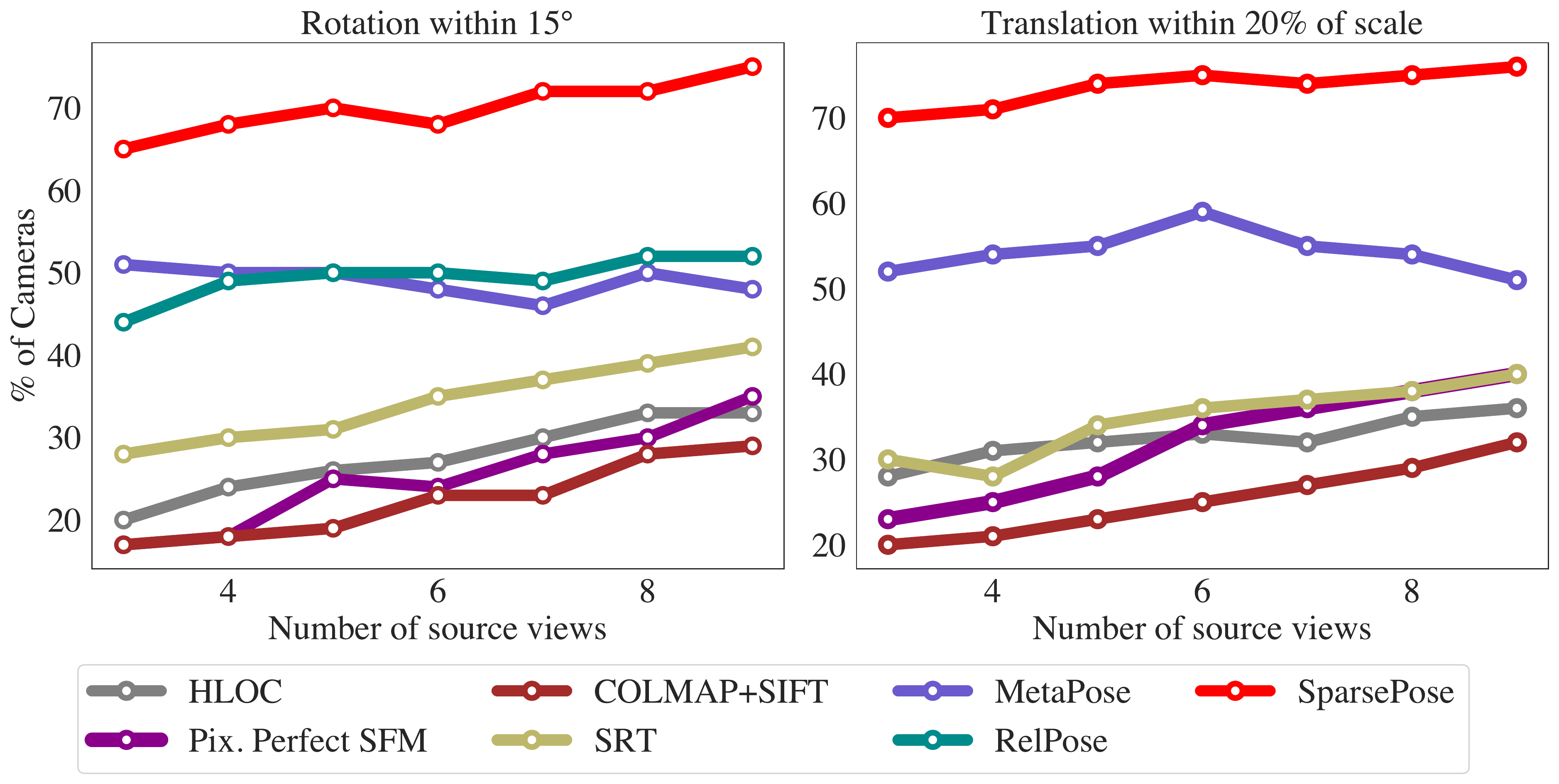}
    \caption{\textbf{Quantitative evaluation of sparse-view camera pose estimation.}
    We evaluate the quality of rotations and translations for varying numbers of source views. We show the percentage of cameras that were predicted to be within $15^\circ$ of the ground truth (left) and translations that were predicted within 20\% of the scale of the scene compared to ground truth (right). \methodNameAbbrev outperforms both classical and learning-based methods.}
    \label{fig:rot-trans-plot}
\end{figure}

\begin{figure}[t!]
    \centering
    \includegraphics[width=\linewidth]{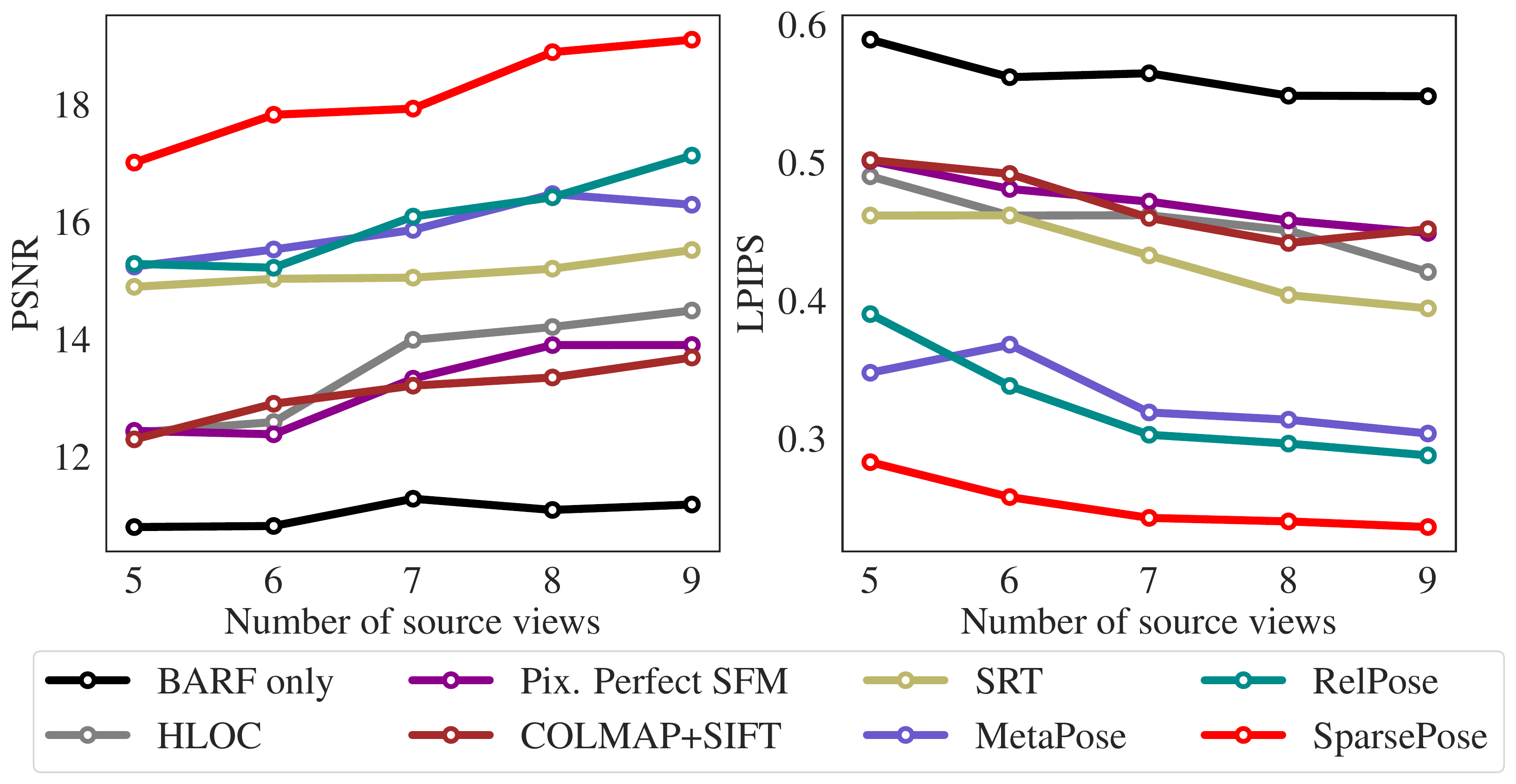}
    \caption{
    \textbf{Quantitative evaluation of sparse-view, novel view synthesis.} We use the camera poses predicted by each method to perform novel view synthesis; \methodNameAbbrev significantly outperforms other baseline methods for this task in terms of PSNR (higher is better) and LPIPS (lower is better).
    }
    \label{fig:recon-plot}
\end{figure}

\begin{figure*}[t!]
    \centering
    \includegraphics[width=\linewidth]{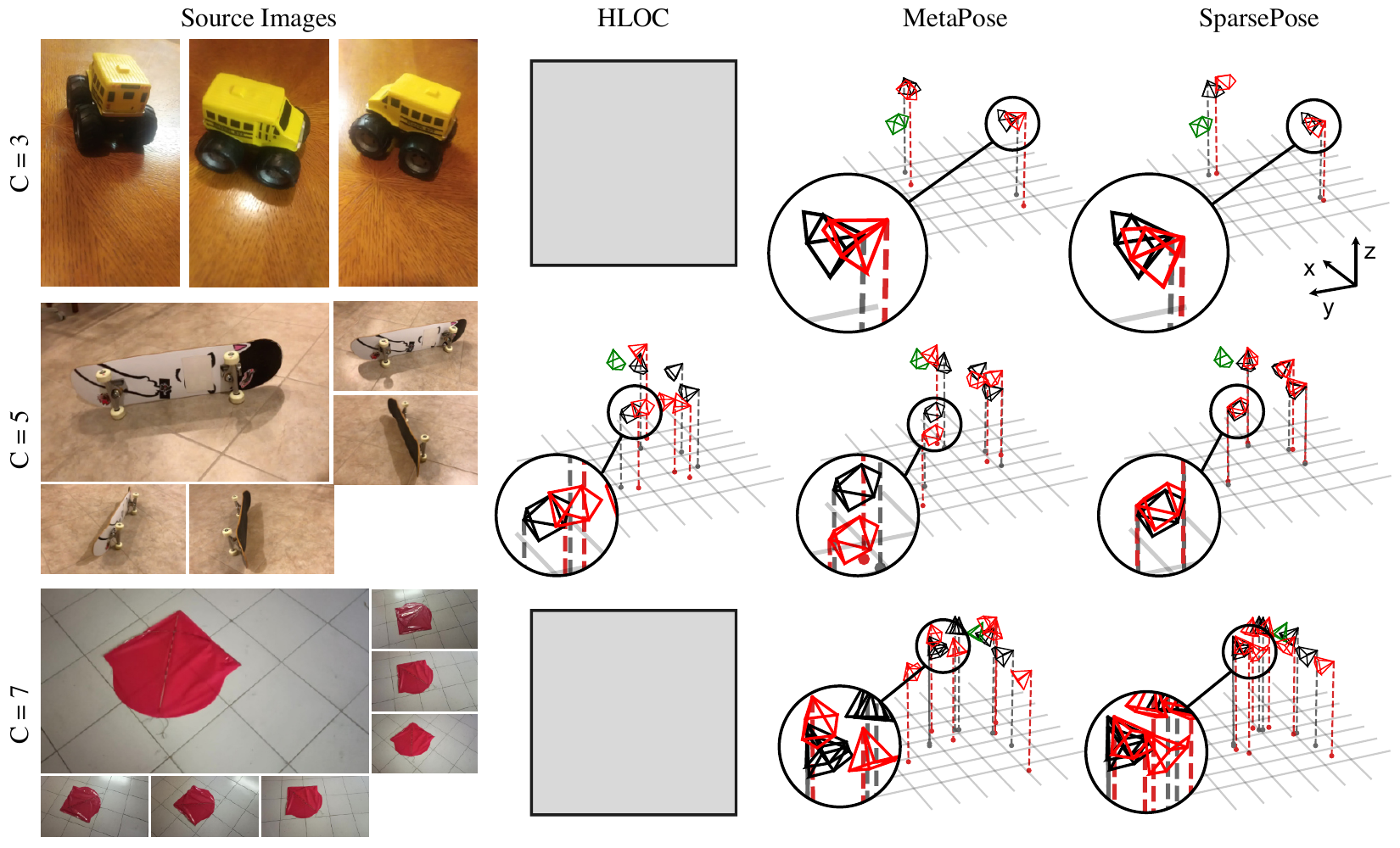}
    \caption{
    \textbf{Visualization of camera poses.}
    We compare the predicted camera poses for various methods and different number of images, camera centers are projected to the $x-y$ plane for comparison. The ground truth poses are shown in black, predicted poses in \textcolor{red}{red}, and the first camera for each sequence (used to align predictions) in \textcolor{green}{green}.  Gray boxes indicate failure to converge.
    }
    \label{fig:qual-cameras}
\end{figure*}

\begin{figure*}[t!]
    \centering
    \includegraphics[width=\linewidth]{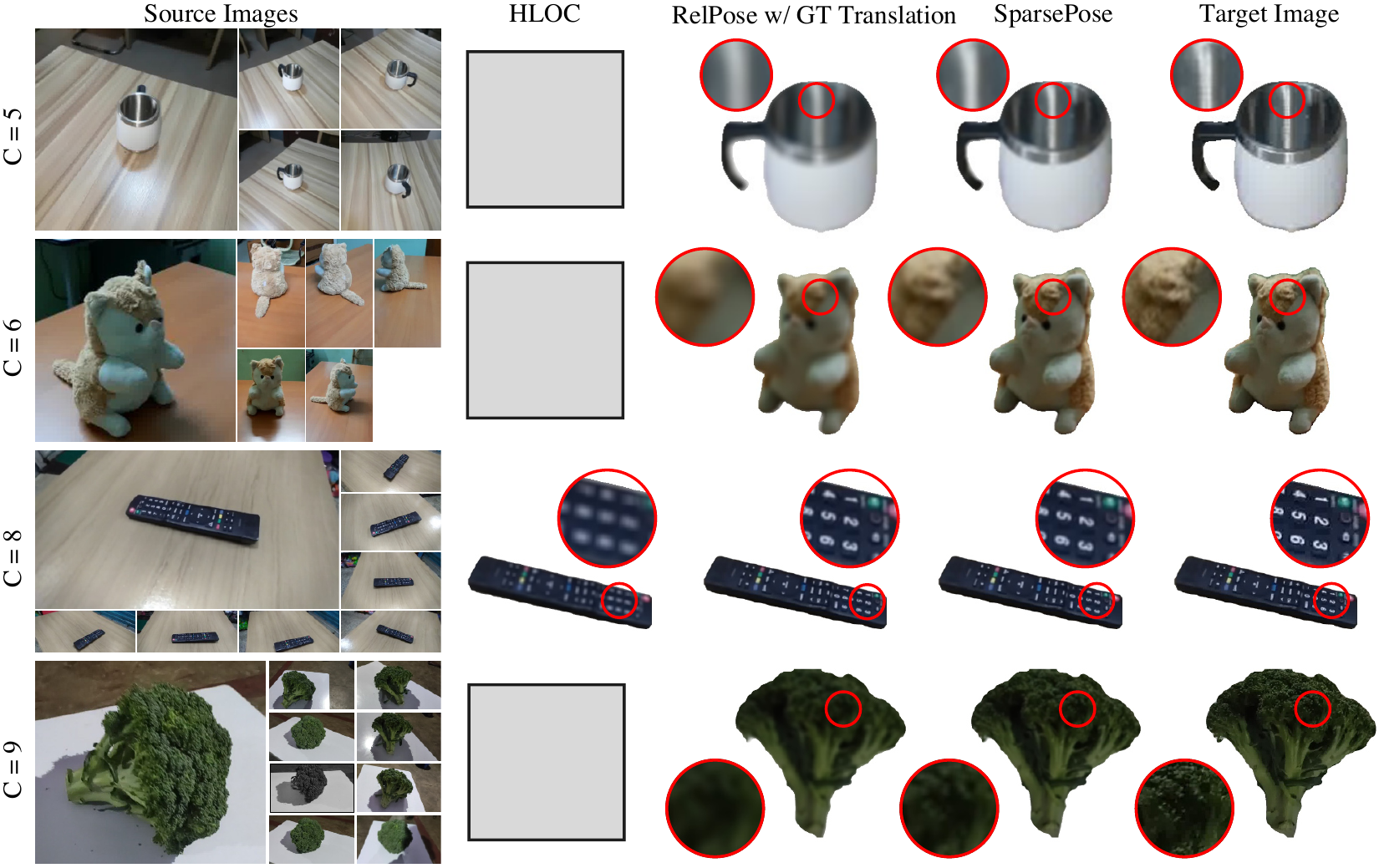}
    \caption{
    \textbf{Visualizing the rendered from few-sparse unposed images.}
    We use the initial predicted poses by each of the methods, and refine them using BARF \cite{barf}, and a category-centric pretrained NeRFormer model, 
    trained on the category.
    The importance of predicting accurate initial poses can be seen since \methodNameAbbrev is able to generate photorealistic renders. Gray boxes indicate failure to converge.
    }
    \label{fig:qual-renders}
\end{figure*}

\begin{figure}[t!]
    \centering
    \includegraphics[width=\linewidth]{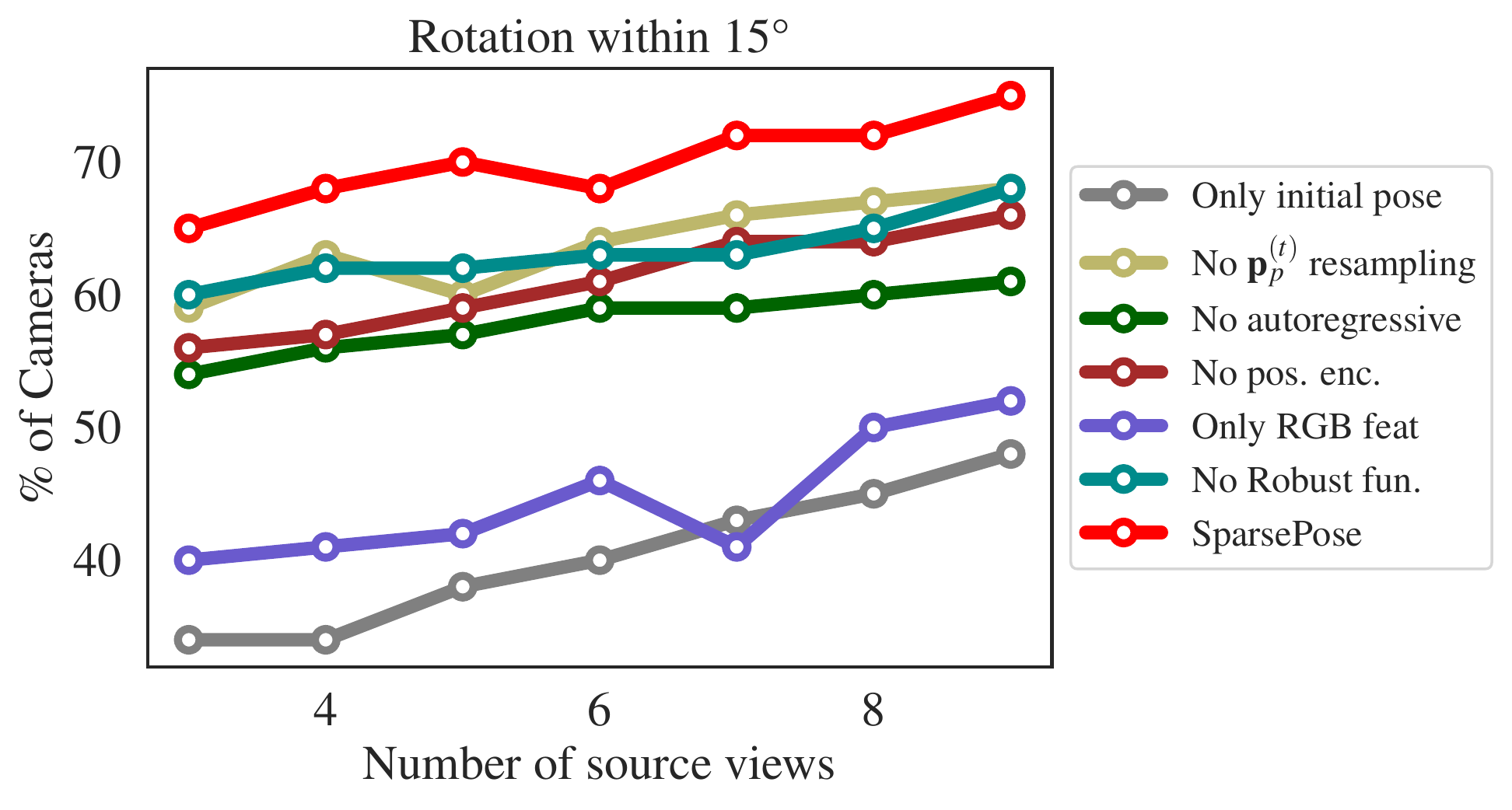}
    \caption{\textbf{Ablation results over different design choices for \methodNameAbbrev}.
    We perform an exhaustive ablation to justify the design choices made in the paper, and report the ability of the model to correctly predict the rotations
    within $15^\circ$ of the ground truth rotations, over different number of source views.
    }
    \label{fig:ablation}
\end{figure}

\section{Experiments}
\label{sec:exps}
For the task of sparse-view camera pose estimation, we consider both classical Structure-from-Motion baselines and modern deep learning based techniques.
More specifically, we compare \methodNameAbbrev against three classical SfM baselines: 
\begin{itemize}
    \item COLMAP with SIFT features \cite{sfm_revisited,sift};
    \item Hierarchical Localization (HLOC) \cite{hloc} which uses COLMAP with SuperPoint for feature extraction \cite{superpoint} and SuperGlue for image matching \cite{superglue};
    \item Pixel-Perfect SfM~\cite{pixelperfectsfm} which is a state-of-the-art SfM method that refines COLMAP camera poses using ``featuremetric bundle adjustment''.
\end{itemize}
We further compare \methodNameAbbrev against:
\begin{itemize}
    \item Scene Representation Transformer (SRT) \cite{srt} by adding an additional layer to the transformer output which jointly learns 3D reconstruction and pose estimation over the large dataset; 
    \item MetaPose \cite{metapose} where we initialize the camera estimates using our initial pose estimation model, and perform pose refinement using their architecture;
    \item RelPose \cite{relpose} which only predicts rotations by learning a energy-based probabilistic model over $\mathrm{SO}(3)$, given a set of images by only considering the \textit{global} features across the images in the scene.
\end{itemize}

For camera pose estimation, since the \textit{ground-truth} cameras from CO3D \cite{co3d} are in arbirtary coordinate frames, we measure the 
relative rotations and translations. 
That is, we measure the absolute angle difference between the ground truth and the predicted camera poses by using the Rodriguez's formula between 
the rotations \cite{rodrigues_formula}, and measure the $\ell$-2 norm between the translations.
Following RelPose \cite{relpose}, we report the percentage of cameras that were predicted within $15^\circ$ of the ground truth rotation.
For translations, since the scale of the scene changes between sequences, we report the percentage of cameras that were within $20\%$ of the scale of the scene.

We then evaluate the predicted cameras for a downstream task of few-view 3D reconstruction on 20 unseen test categories.
For the novel-view synthesis task, we report the Peak-Signal-to-Noise-Ratio (PSNR) which measures the difference in the RGB space, 
and Learned Perceptual Image Patch Similarity (LPIPS) which measures the difference as a ``perceptual'' score.

\subsection{Wide-baseline camera pose estimation}
We report quantitative results with different numbers of source views in Figure \ref{fig:rot-trans-plot} which shows the percentage of predicted rotations within $15^\circ$ of ground truth and predicted translations within $20\%$ of the scale of the scene.
The ground truth is obtained using SfM on dense videos with more than $300$ images~\cite{sfm_revisited}.
\methodNameAbbrev is able to significantly outperform both classical SfM and learning-based baselines by a significant margin.
Moreover, \methodNameAbbrev consistently improves in performance as the number of source views increases, which is not the case across all baselines (e.g., MetaPose~\cite{metapose}).
Using our method, $65-80\%$ of predicted camera locations and orientations fall within the thresholds described above. 
In contrast, correspondence based approaches such as COLMAP~\cite{sfm_revisited, sift}, HLOC~\cite{hloc}, and 
Pixel-Perfect SfM~\cite{pixelperfectsfm}, are only able to recover $20-40\%$ of the cameras within the thresholds for rotation and translation with $C{<}10$.
This significant difference in performance motivates the effectiveness of learning appearance priors and learning to perform geometry-based pose refinement.
We note that RelPose \cite{relpose} only predicts rotations, and therefore cannot be evaluated on the translation prediction task.

We also show a visualization of the predicted and ground truth cameras in Figure~\ref{fig:qual-cameras}, where we project the camera centers onto the $x$--$y$ plane to help visualize 3D offsets in the camera centers.
\methodNameAbbrev predicts accurate camera poses given a sparse set of images with very wide baselines, significantly outperforming other methods.
Even on very challenging sequences with uneven lighting and low textural information (e.g., $C{=}7$), our method predicts accurate cameras, which is important in practical cases.
We note that on many sequences HLOC fails to register all the source images and so does not converge to a usable output; we cannot include a result in this case.

\subsection{Sparse-view 3D reconstruction}
We test the predicted cameras on the downstream task of sparse-view 3D reconstruction using a NeRFormer that is trained on the test categories \cite{co3d} (but not any of the test \textit{sequences} we evaluate on).
Training is performed using the default hyperparameters from PyTorch3D~\cite{pytorch3d}.
During evaluation, we further finetune the cameras using BARF~\cite{barf}, an off-the-shelf 3D reconstruction and pose refinement technique which updates the camera poses by minimizing the photometric loss.
In addition to the previous baselines, we add a ``BARF-only'' baseline, which performs refinement after a unit camera initialization for finetuning with BARF (i.e., $\rot = \mathbb{I}, \mathbf{T} = 0$).

Quantitative results are shown in Figure~\ref{fig:recon-plot}. Here, \methodNameAbbrev significantly outperforms all baselines.
Most importantly, we show that, while predicted rotations and translations are not perfect, we can yet recover high-fidelity 3D reconstructions in the downstream task.
Additionally, we find that BARF without good initial pose estimates does not converge to accurate camera poses.
This further highlights the importance of accurate initial pose estimates.
For the comparison to RelPose, we use the ground truth translations as predicted by CO3D (since the method does not predict translations); yet, \methodNameAbbrev significantly outperforms RelPose across all numbers of the source views in both visual metrics.
Finally, we also show qualitative novel-view synthesis results in Figure \ref{fig:qual-renders}. 
\methodNameAbbrev results in significantly better novel-view synthesis compared to baselines such as RelPose~(with ground truth translations) and HLOC.
Note that when using significantly more source images, the performance of conventional methods such as HLOC~\cite{hloc} or COLMAP~\cite{sfm_revisited} typically improves. 
For example, see the analysis in Zhang et al.~\cite{relpose} which shows performance of such methods with up to 20 images.

\subsection{Ablation study}
We provide an ablation study of \methodNameAbbrev using different variants of the model to justify the design choices~(see Figure~\ref{fig:ablation}): (1) only initial pose (i.e., no pose refinement), (2) no resampling of the 3D points between iterations, (3) using an MLP instead of \textsc{lstm}, (4) no positional encoding on the inputs to $\mathcal{T}_\text{refine}$, (5) using RGB values instead of features from $\mathcal{E}_{\text{init}}$ for the refinement step, (6) no robust kernel \cite{robust_l2}. 
We show that \methodNameAbbrev with the proposed design outperforms the other variants.
Interestingly, using the initial poses $\mathbf{R}^{(0)}, \mathbf{t}^{(0)}$, we achieve performance similar to classical SfM methods, which highlights the importance of learned appearance priors from large datasets.
The best performance is achieved when refining camera poses using the proposed method, including positional encoding, auto-regressive prediction, etc. 

\section{Conclusion}
\label{sec:conclusion}

In this paper, we presented \methodNameAbbrev, a learning-based solution to perform sparse-view camera pose estimation from wide-baseline input images.
The strong performance of the method highlights the utility of leveraging large object-centric datasets for learning pose regression and refinement.
Moreover, we show that accurate few-view pose estimation can enable few-view novel-view synthesis even from challenging ``in-the-wild'' datasets, where our method outperforms other baselines.

There remain many potential directions for future work, among which, we believe that joint methods for pose regression and 3D scene geometry prediction may enable further improved capabilities for novel view synthesis from sparse views. 
Additional work on learning where and \textit{how} to sample the 3D points used in our pose refinement step may help extend the approach to other camera motions beyond the ``tracked-to-object'' camera poses common in the CO3D dataset.
Furthermore, it may be interesting to apply variants of our approach to the challenge of non-rigid structure from motion, 
where correspondence-based methods tend to fail.
Learning a prior over \textit{motion}, may help with camera pose estimation for highly non-rigid scenes, such as for scenes with smoke, loose clothing, or humans performing complex movements.
Finally, our work may be broadly relevant to improving the robustness of robotic vision, autonomous navigation systems, and for efficient digital asset creation. 
We envision that creators will be able to take a few sparse images of common objects and generate photorealistic 3D assets for applications in augmented or virtual reality.




\paragraph{Acknowledgements} This project was supported in part by NSERC under the RGPIN program.


{\small
\bibliographystyle{ieee_fullname}
\bibliography{macros,main}

\begin{thebibliography}{10}\itemsep=-1pt

\bibitem{large_bundle_adjustment}
Sameer Agarwal, Noah Snavely, Steven~M Seitz, and Richard Szeliski.
\newblock Bundle adjustment in the large.
\newblock In {\em ECCV}, 2010.

\bibitem{learning2learn}
Marcin Andrychowicz, Misha Denil, Sergio Gomez, Matthew~W Hoffman, David Pfau,
  Tom Schaul, Brendan Shillingford, and Nando De~Freitas.
\newblock Learning to learn by gradient descent by gradient descent.
\newblock {\em NeurIPS}, 2016.

\bibitem{rootsift}
Relja Arandjelovi{\'c} and Andrew Zisserman.
\newblock Three things everyone should know to improve object retrieval.
\newblock In {\em CVPR}, 2012.

\bibitem{robust_l2}
Jonathan~T Barron.
\newblock A general and adaptive robust loss function.
\newblock In {\em CVPR}, 2019.

\bibitem{mipnerf360}
Jonathan~T Barron, Ben Mildenhall, Dor Verbin, Pratul~P Srinivasan, and Peter
  Hedman.
\newblock Mip-nerf 360: Unbounded anti-aliased neural radiance fields.
\newblock In {\em CVPR}, 2022.

\bibitem{bay2008speeded}
Herbert Bay, Andreas Ess, Tinne Tuytelaars, and Luc Van~Gool.
\newblock Speeded-up robust features (surf).
\newblock {\em Computer vision and image understanding}, 2008.

\bibitem{boss2022samurai}
Mark Boss, Andreas Engelhardt, Abhishek Kar, Yuanzhen Li, Deqing Sun,
  Jonathan~T Barron, Hendrik Lensch, and Varun Jampani.
\newblock Samurai: Shape and material from unconstrained real-world arbitrary
  image collections.
\newblock {\em NeurIPS}, 2022.

\bibitem{dino}
Mathilde Caron, Hugo Touvron, Ishan Misra, Herv{\'e} J{\'e}gou, Julien Mairal,
  Piotr Bojanowski, and Armand Joulin.
\newblock Emerging properties in self-supervised vision transformers.
\newblock In {\em CVPR}, 2021.

\bibitem{survey_rigid_pose}
Jiale Chen, Lijun Zhang, Yi Liu, and Chi Xu.
\newblock Survey on 6d pose estimation of rigid object.
\newblock In {\em Chinese Control Conference (CCC)}, 2020.

\bibitem{dirnet}
Kefan Chen, Noah Snavely, and Ameesh Makadia.
\newblock Wide-baseline relative camera pose estimation with directional
  learning.
\newblock In {\em CVPR}, 2021.

\bibitem{nsm}
Yun-Chun Chen, Haoda Li, Dylan Turpin, Alec Jacobson, and Animesh Garg.
\newblock Neural shape mating: Self-supervised object assembly with adversarial
  shape priors.
\newblock In {\em CVPR}, 2022.

\bibitem{garf}
Shin-Fang Chng, Sameera Ramasinghe, Jamie Sherrah, and Simon Lucey.
\newblock Garf: Gaussian activated radiance fields for high fidelity
  reconstruction and pose estimation.
\newblock {\em arXiv e-prints}, 2022.

\bibitem{dsnerf}
Kangle Deng, Andrew Liu, Jun-Yan Zhu, and Deva Ramanan.
\newblock Depth-supervised nerf: Fewer views and faster training for free.
\newblock In {\em CVPR}, 2022.

\bibitem{superpoint}
Daniel DeTone, Tomasz Malisiewicz, and Andrew Rabinovich.
\newblock Superpoint: Self-supervised interest point detection and description.
\newblock In {\em CVPR (workshops)}, 2018.

\bibitem{vit}
Alexey Dosovitskiy, Lucas Beyer, Alexander Kolesnikov, Dirk Weissenborn,
  Xiaohua Zhai, Thomas Unterthiner, Mostafa Dehghani, Matthias Minderer, Georg
  Heigold, Sylvain Gelly, et~al.
\newblock An image is worth 16x16 words: Transformers for image recognition at
  scale.
\newblock {\em arXiv preprint arXiv:2010.11929}, 2020.

\bibitem{enr}
Emilien Dupont, Miguel~Bautista Martin, Alex Colburn, Aditya Sankar, Josh
  Susskind, and Qi Shan.
\newblock Equivariant neural rendering.
\newblock In {\em International Conference on Machine Learning}. PMLR, 2020.

\bibitem{slam}
Hugh Durrant-Whyte and Tim Bailey.
\newblock Simultaneous localization and mapping: part i.
\newblock {\em IEEE robotics \& automation magazine}, 2006.

\bibitem{ucmr}
Shubham Goel, Angjoo Kanazawa, and Jitendra Malik.
\newblock Shape and viewpoint without keypoints.
\newblock In {\em ECCV}, 2020.

\bibitem{nerf2nerf}
Lily Goli, Daniel Rebain, Sara Sabour, Animesh Garg, and Andrea Tagliasacchi.
\newblock {nerf2nerf}: pairwise registration of neural radiance fields.
\newblock {\em arXiv preprint arXiv:2211.01600}, 2022.

\bibitem{hartley2003multiple}
Richard Hartley and Andrew Zisserman.
\newblock {\em Multiple view geometry in computer vision}.
\newblock Cambridge university press, 2003.

\bibitem{wce}
Philipp Henzler, Jeremy Reizenstein, Patrick Labatut, Roman Shapovalov, Tobias
  Ritschel, Andrea Vedaldi, and David Novotny.
\newblock Unsupervised learning of 3d object categories from videos in the
  wild.
\newblock In {\em CVPR}, 2021.

\bibitem{lstm}
Sepp Hochreiter and J{\"u}rgen Schmidhuber.
\newblock Long short-term memory.
\newblock {\em Neural computation}, 9(8), 1997.

\bibitem{jain2021putting}
Ajay Jain, Matthew Tancik, and Pieter Abbeel.
\newblock Putting nerf on a diet: Semantically consistent few-shot view
  synthesis.
\newblock In {\em ICCV}, 2021.

\bibitem{codenerf}
Wonbong Jang and Lourdes Agapito.
\newblock Codenerf: Disentangled neural radiance fields for object categories.
\newblock In {\em ICCV}, 2021.

\bibitem{scnerf}
Yoonwoo Jeong, Seokjun Ahn, Christopher Choy, Anima Anandkumar, Minsu Cho, and
  Jaesik Park.
\newblock Self-calibrating neural radiance fields.
\newblock In {\em ICCV}, 2021.

\bibitem{jin2021image}
Yuhe Jin, Dmytro Mishkin, Anastasiia Mishchuk, Jiri Matas, Pascal Fua,
  Kwang~Moo Yi, and Eduard Trulls.
\newblock Image matching across wide baselines: From paper to practice.
\newblock {\em IJCV}, 129(2), 2021.

\bibitem{joo2015panoptic}
Hanbyul Joo and Hao Liu.
\newblock Panoptic studio: A massively multiview system for social motion
  capture.
\newblock In {\em ICCV}, 2015.

\bibitem{ssd-6d}
Wadim Kehl, Fabian Manhardt, Federico Tombari, Slobodan Ilic, and Nassir Navab.
\newblock Ssd-6d: Making rgb-based 3d detection and 6d pose estimation great
  again.
\newblock In {\em ICCV}, 2017.

\bibitem{posenet}
Alex Kendall, Matthew Grimes, and Roberto Cipolla.
\newblock Posenet: A convolutional network for real-time 6-dof camera
  relocalization.
\newblock In {\em ICCV}, 2015.

\bibitem{adam}
Diederik~P Kingma and Jimmy Ba.
\newblock Adam: A method for stochastic optimization.
\newblock {\em arXiv preprint arXiv:1412.6980}, 2014.

\bibitem{ttp}
Filippos Kokkinos and Iasonas Kokkinos.
\newblock To the point: Correspondence-driven monocular 3d category
  reconstruction.
\newblock {\em NeurIPS}, 2021.

\bibitem{csm}
Nilesh Kulkarni, Abhinav Gupta, and Shubham Tulsiani.
\newblock Canonical surface mapping via geometric cycle consistency.
\newblock In {\em ICCV}, 2019.

\bibitem{umr}
Xueting Li, Sifei Liu, Kihwan Kim, Shalini~De Mello, Varun Jampani, Ming-Hsuan
  Yang, and Jan Kautz.
\newblock Self-supervised single-view 3d reconstruction via semantic
  consistency.
\newblock In {\em ECCV}, 2020.

\bibitem{barf}
Chen-Hsuan Lin, Wei-Chiu Ma, Antonio Torralba, and Simon Lucey.
\newblock Barf: Bundle-adjusting neural radiance fields.
\newblock In {\em ICCV}, 2021.

\bibitem{lindell2021autoint}
David~B Lindell, Julien~NP Martel, and Gordon Wetzstein.
\newblock {AutoInt}: Automatic integration for fast neural volume rendering.
\newblock In {\em CVPR}, 2021.

\bibitem{lindell2022bacon}
David~B Lindell, Dave Van~Veen, Jeong~Joon Park, and Gordon Wetzstein.
\newblock {BACON}: Band-limited coordinate networks for multiscale scene
  representation.
\newblock In {\em CVPR}, 2022.

\bibitem{pixelperfectsfm}
Philipp Lindenberger, Paul-Edouard Sarlin, Viktor Larsson, and Marc Pollefeys.
\newblock Pixel-perfect structure-from-motion with featuremetric refinement.
\newblock In {\em ICCV}, 2021.

\bibitem{siftflow}
Ce Liu, Jenny Yuen, and Antonio Torralba.
\newblock Sift flow: Dense correspondence across scenes and its applications.
\newblock {\em IEEE TPAMI}, 2010.

\bibitem{sift}
David~G Lowe.
\newblock Distinctive image features from scale-invariant keypoints.
\newblock {\em IJCV}, 2004.

\bibitem{vircoor}
Wei-Chiu Ma, Anqi~Joyce Yang, Shenlong Wang, Raquel Urtasun, and Antonio
  Torralba.
\newblock Virtual correspondence: Humans as a cue for extreme-view geometry.
\newblock In {\em CVPR}, 2022.

\bibitem{nerf}
Ben Mildenhall, Pratul~P. Srinivasan, Matthew Tancik, Jonathan~T. Barron, Ravi
  Ramamoorthi, and Ren Ng.
\newblock {NeRF: Representing Scenes as Neural Radiance Fields for View
  Synthesis}.
\newblock In {\em ECCV}, 2020.

\bibitem{moravec1980obstacle}
Hans~Peter Moravec.
\newblock {\em Obstacle avoidance and navigation in the real world by a seeing
  robot rover}.
\newblock Stanford University, 1980.

\bibitem{niemeyer2022regnerf}
Michael Niemeyer, Jonathan~T Barron, Ben Mildenhall, Mehdi~SM Sajjadi, Andreas
  Geiger, and Noha Radwan.
\newblock Regnerf: Regularizing neural radiance fields for view synthesis from
  sparse inputs.
\newblock In {\em CVPR}, 2022.

\bibitem{nister2004visual}
David Nist{\'e}r, Oleg Naroditsky, and James Bergen.
\newblock Visual odometry.
\newblock In {\em CVPR}, 2004.

\bibitem{c3dpo}
David Novotny, Nikhila Ravi, Benjamin Graham, Natalia Neverova, and Andrea
  Vedaldi.
\newblock C3dpo: Canonical 3d pose networks for non-rigid structure from
  motion.
\newblock In {\em ICCV}, 2019.

\bibitem{sfm_survey}
Onur {\"O}zye{\c{s}}il, Vladislav Voroninski, Ronen Basri, and Amit Singer.
\newblock A survey of structure from motion*.
\newblock {\em Acta Numerica}, 2017.

\bibitem{ozyecsil2017survey}
Onur {\"O}zye{\c{s}}il, Vladislav Voroninski, Ronen Basri, and Amit Singer.
\newblock A survey of structure from motion*.
\newblock {\em Acta Numerica}, 2017.

\bibitem{paszke2019pytorch}
Adam Paszke, Sam Gross, Francisco Massa, Adam Lerer, James Bradbury, et~al.
\newblock Pytorch: An imperative style, high-performance deep learning library.
\newblock In {\em NeurIPS}, 2019.

\bibitem{qi2017pointnet}
Charles~R. Qi, Hao Su, Kaichun Mo, and Leonidas~J. Guibas.
\newblock Pointnet: Deep learning on point sets for {3D} classification and
  segmentation.
\newblock In {\em CVPR}, 2017.

\bibitem{pytorch3d}
Nikhila Ravi, Jeremy Reizenstein, David Novotny, Taylor Gordon, Wan-Yen Lo,
  Justin Johnson, and Georgia Gkioxari.
\newblock Accelerating 3d deep learning with pytorch3d.
\newblock {\em arXiv preprint arXiv:2007.08501}, 2020.

\bibitem{lstm_fewshot}
Sachin Ravi and Hugo Larochelle.
\newblock Optimization as a model for few-shot learning.
\newblock In {\em ICLR}, 2017.

\bibitem{co3d}
Jeremy Reizenstein, Roman Shapovalov, Philipp Henzler, Luca Sbordone, Patrick
  Labatut, and David Novotny.
\newblock Common objects in 3d: Large-scale learning and evaluation of
  real-life 3d category reconstruction.
\newblock In {\em ICCV}, 2021.

\bibitem{rodrigues_formula}
Olinde Rodrigues.
\newblock Des lois g{\'e}om{\'e}triques qui r{\'e}gissent les d{\'e}placements
  d’un syst{\`e}me solide dans l’espace, et de la variation des
  coordonn{\'e}es provenant de ces d{\'e}placements consid{\'e}r{\'e}s
  ind{\'e}pendamment des causes qui peuvent les produire.
\newblock {\em J. Math. Pures Appl}, 1840.

\bibitem{requires_depth_multiview}
Barbara Roessle and Matthias Nie{\ss}ner.
\newblock End2end multi-view feature matching using differentiable pose
  optimization.
\newblock {\em arXiv preprint arXiv:2205.01694}, 2022.

\bibitem{srt}
Mehdi~SM Sajjadi, Henning Meyer, Etienne Pot, Urs Bergmann, Klaus Greff, Noha
  Radwan, Suhani Vora, Mario Lu{\v{c}}i{\'c}, Daniel Duckworth, Alexey
  Dosovitskiy, et~al.
\newblock Scene representation transformer: Geometry-free novel view synthesis
  through set-latent scene representations.
\newblock In {\em CVPR}, 2022.

\bibitem{hloc}
Paul-Edouard Sarlin, Cesar Cadena, Roland Siegwart, and Marcin Dymczyk.
\newblock From coarse to fine: Robust hierarchical localization at large scale.
\newblock In {\em CVPR}, 2019.

\bibitem{superglue}
Paul-Edouard Sarlin, Daniel DeTone, Tomasz Malisiewicz, and Andrew Rabinovich.
\newblock Superglue: Learning feature matching with graph neural networks.
\newblock In {\em CVPR}, 2020.

\bibitem{sfm_revisited}
Johannes~L Schonberger and Jan-Michael Frahm.
\newblock Structure-from-motion revisited.
\newblock In {\em CVPR}, 2016.

\bibitem{anerf}
Shih-Yang Su, Frank Yu, Michael Zollh{\"o}fer, and Helge Rhodin.
\newblock A-nerf: Articulated neural radiance fields for learning human shape,
  appearance, and pose.
\newblock {\em NeurIPS}, 2021.

\bibitem{loftr}
Jiaming Sun, Zehong Shen, Yuang Wang, Hujun Bao, and Xiaowei Zhou.
\newblock Loftr: Detector-free local feature matching with transformers.
\newblock In {\em CVPR}, 2021.

\bibitem{blocknerf}
Matthew Tancik, Vincent Casser, Xinchen Yan, Sabeek Pradhan, Ben Mildenhall,
  Pratul~P Srinivasan, Jonathan~T Barron, and Henrik Kretzschmar.
\newblock Block-nerf: Scalable large scene neural view synthesis.
\newblock In {\em CVPR}, 2022.

\bibitem{bundle_adjustment}
Bill Triggs, Philip~F McLauchlan, Richard~I Hartley, and Andrew~W Fitzgibbon.
\newblock Bundle adjustment—a modern synthesis.
\newblock In {\em International workshop on vision algorithms}, 1999.

\bibitem{metapose}
Ben Usman, Andrea Tagliasacchi, Kate Saenko, and Avneesh Sud.
\newblock Metapose: Fast 3d pose from multiple views without 3d supervision.
\newblock In {\em CVPR}, 2022.

\bibitem{attention}
Ashish Vaswani, Noam Shazeer, Niki Parmar, Jakob Uszkoreit, Llion Jones,
  Aidan~N Gomez, {\L}ukasz Kaiser, and Illia Polosukhin.
\newblock Attention is all you need.
\newblock {\em NeurIPS}, 30, 2017.

\bibitem{deep_vo3}
Sen Wang, Ronald Clark, Hongkai Wen, and Niki Trigoni.
\newblock Deepvo: Towards end-to-end visual odometry with deep recurrent
  convolutional neural networks.
\newblock In {\em Proc. {ICRA}}. IEEE, 2017.

\bibitem{nerfmm}
Zirui Wang, Shangzhe Wu, Weidi Xie, Min Chen, and Victor~Adrian Prisacariu.
\newblock Nerf--: Neural radiance fields without known camera parameters.
\newblock {\em arXiv preprint arXiv:2102.07064}, 2021.

\bibitem{threedim}
Daniel Watson, William Chan, Ricardo Martin-Brualla, Jonathan Ho, Andrea
  Tagliasacchi, and Mohammad Norouzi.
\newblock Novel view synthesis with diffusion models.
\newblock {\em arXiv preprint arXiv:2210.04628}, 2022.

\bibitem{dove}
Shangzhe Wu, Tomas Jakab, Christian Rupprecht, and Andrea Vedaldi.
\newblock Dove: Learning deformable 3d objects by watching videos.
\newblock {\em arXiv preprint arXiv:2107.10844}, 2021.

\bibitem{unsup3d}
Shangzhe Wu, Christian Rupprecht, and Andrea Vedaldi.
\newblock Unsupervised learning of probably symmetric deformable 3d objects
  from images in the wild.
\newblock In {\em CVPR}, 2020.

\bibitem{object_pose_2}
Yu Xiang, Tanner Schmidt, Venkatraman Narayanan, and Dieter Fox.
\newblock Posecnn: A convolutional neural network for 6d object pose estimation
  in cluttered scenes.
\newblock {\em arXiv preprint arXiv:1711.00199}, 2017.

\bibitem{lasr}
Gengshan Yang, Deqing Sun, Varun Jampani, Daniel Vlasic, Forrester Cole, Huiwen
  Chang, Deva Ramanan, William~T Freeman, and Ce Liu.
\newblock Lasr: Learning articulated shape reconstruction from a monocular
  video.
\newblock In {\em CVPR}, 2021.

\bibitem{deep_vo2}
Nan Yang, Lukas~von Stumberg, Rui Wang, and Daniel Cremers.
\newblock D3vo: Deep depth, deep pose and deep uncertainty for monocular visual
  odometry.
\newblock In {\em CVPR}, 2020.

\bibitem{inerf}
Lin Yen-Chen, Pete Florence, Jonathan~T Barron, Alberto Rodriguez, Phillip
  Isola, and Tsung-Yi Lin.
\newblock inerf: Inverting neural radiance fields for pose estimation.
\newblock In {\em {IROS}}, 2021.

\bibitem{lift}
Kwang~Moo Yi, Eduard Trulls, Vincent Lepetit, and Pascal Fua.
\newblock Lift: Learned invariant feature transform.
\newblock In {\em ECCV}. Springer, 2016.

\bibitem{Yu2020pixelnerf}
Alex Yu, Vickie Ye, Matthew Tancik, and Angjoo Kanazawa.
\newblock {pixelNeRF}: Neural radiance fields from one or few images.
\newblock In {\em CVPR}, 2021.

\bibitem{pixelnerf}
Alex Yu, Vickie Ye, Matthew Tancik, and Angjoo Kanazawa.
\newblock pixelnerf: Neural radiance fields from one or few images.
\newblock In {\em CVPR}, 2021.

\bibitem{humbi_dataset}
Zhixuan Yu, Jae~Shin Yoon, In~Kyu Lee, Prashanth Venkatesh, Jaesik Park, Jihun
  Yu, and Hyun~Soo Park.
\newblock Humbi: A large multiview dataset of human body expressions.
\newblock In {\em CVPR}, 2020.

\bibitem{relpose}
Jason~Y Zhang, Deva Ramanan, and Shubham Tulsiani.
\newblock Relpose: Predicting probabilistic relative rotation for single
  objects in the wild.
\newblock {\em arXiv preprint arXiv:2208.05963}, 2022.

\bibitem{object_pose_1}
Kaifeng Zhang, Yang Fu, Shubhankar Borse, Hong Cai, Fatih Porikli, and Xiaolong
  Wang.
\newblock Self-supervised geometric correspondence for category-level 6d object
  pose estimation in the wild.
\newblock {\em arXiv preprint arXiv:2210.07199}, 2022.

\bibitem{nerf++}
Kai Zhang, Gernot Riegler, Noah Snavely, and Vladlen Koltun.
\newblock Nerf++: Analyzing and improving neural radiance fields.
\newblock {\em arXiv preprint arXiv:2010.07492}, 2020.

\end{thebibliography}
}

\clearpage

\appendix

\begin{strip}    
    \Large
    \noindent
    \centering
    \textbf{
        \noindent
        \Large{SparsePose: Sparse-View Camera Pose Regression and Refinement \\}
        \large{\textit{Suppemental Material}}
    }
\end{strip}

\section{Further Ablation study}

\subsection{LSTM iterations}
\begin{figure}[h!]
    \centering
    \includegraphics[width=\columnwidth]{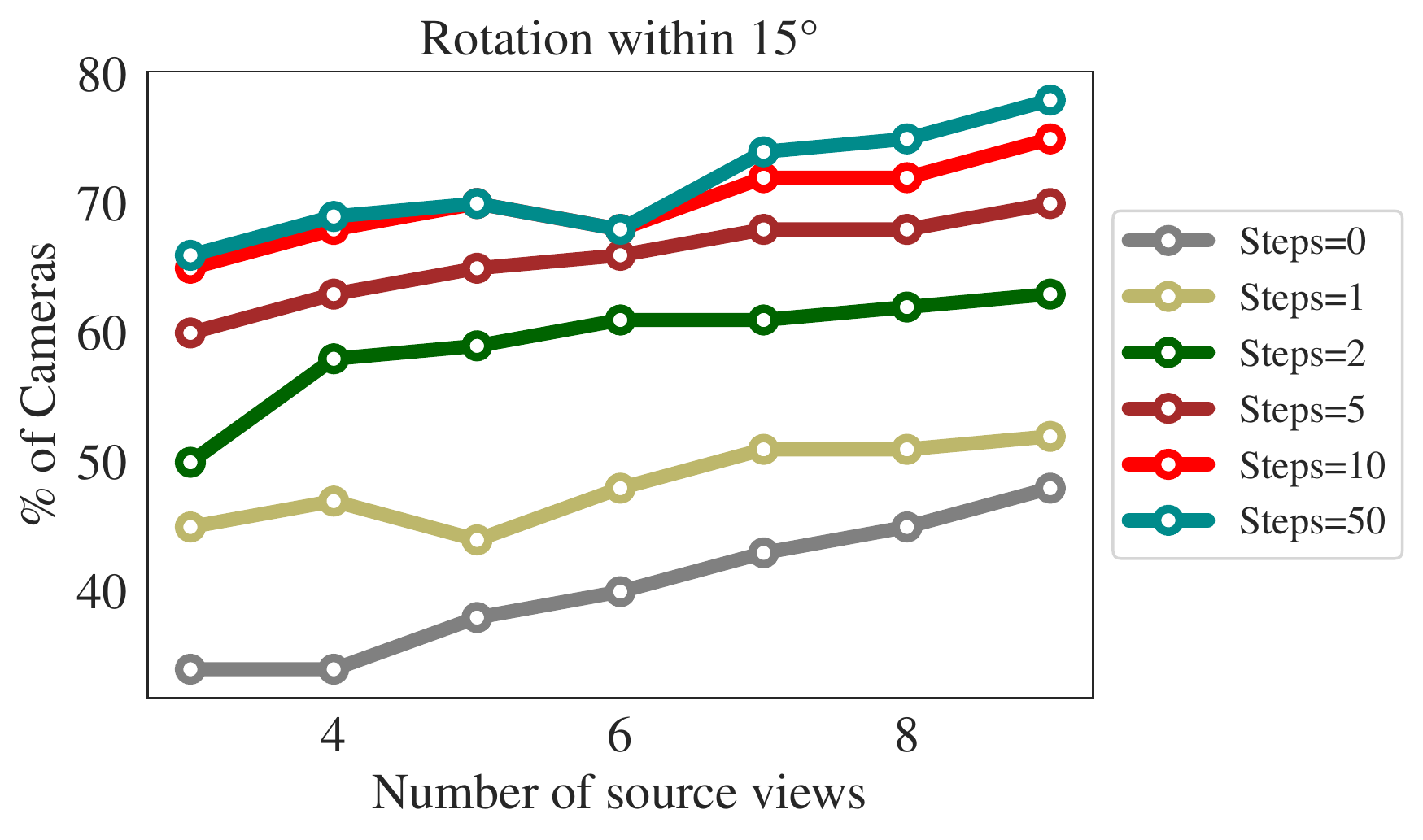}
    \caption{\textbf{Ablation results varying the number of LSTM steps.}}
    \label{fig:lstm-steps}
\end{figure}

We add an additional ablation experiment over the number of steps required for the LSTM.
We vary the number of LSTM iterations between 0 and 50, and report the percentage of cameras that were predicted between $15^\circ$ of ground truth.
We report the results in Figure \ref{fig:lstm-steps}.
As previously noted, all the experiments were performed with 10 LSTM iterations, which balances out speed and accuracy of predictions.
While we observe slight improvements with 50 LSTM iterations, overall, using 10 LSTM iterations performs similarly.
\subsection{Timing Analysis}

\begin{table}[h!]

    \centering
    \begin{tabular}{c|c}
        \toprule
         & Time (seconds) \\
        \midrule
        HLOC \cite{hloc} & 38s  \\
        COLMAP + SIFT \cite{sfm_revisited,sift} & 18s \\ 
        Pix. Perfect SFM \cite{pixelperfectsfm} & 55s \\
        RelPose \cite{relpose} & 48s \\
        SRT \cite{srt} &  \textbf{2.7s} \\
        MetaPose \cite{metapose} & \textbf{2.6s} \\
        \methodNameAbbrev & \textbf{3.6s} \\ 
         \bottomrule
    \end{tabular}
    \caption{
        \textbf{Time (in seconds) to perform registration on a single sequence with 9 source images.}
        To enable fair comparison between all methods, only sequences where \textit{all} baseline methods were able to register all the source images were 
        included in the analysis.
        Each of the methods are run on the same NVIDIA A6000 for fair comparison.
    }
    \label{tab:my_label}

\end{table}

\subsection{Different rotation threshold}

\begin{figure}[h!]
    \centering
    \includegraphics[width=\linewidth]{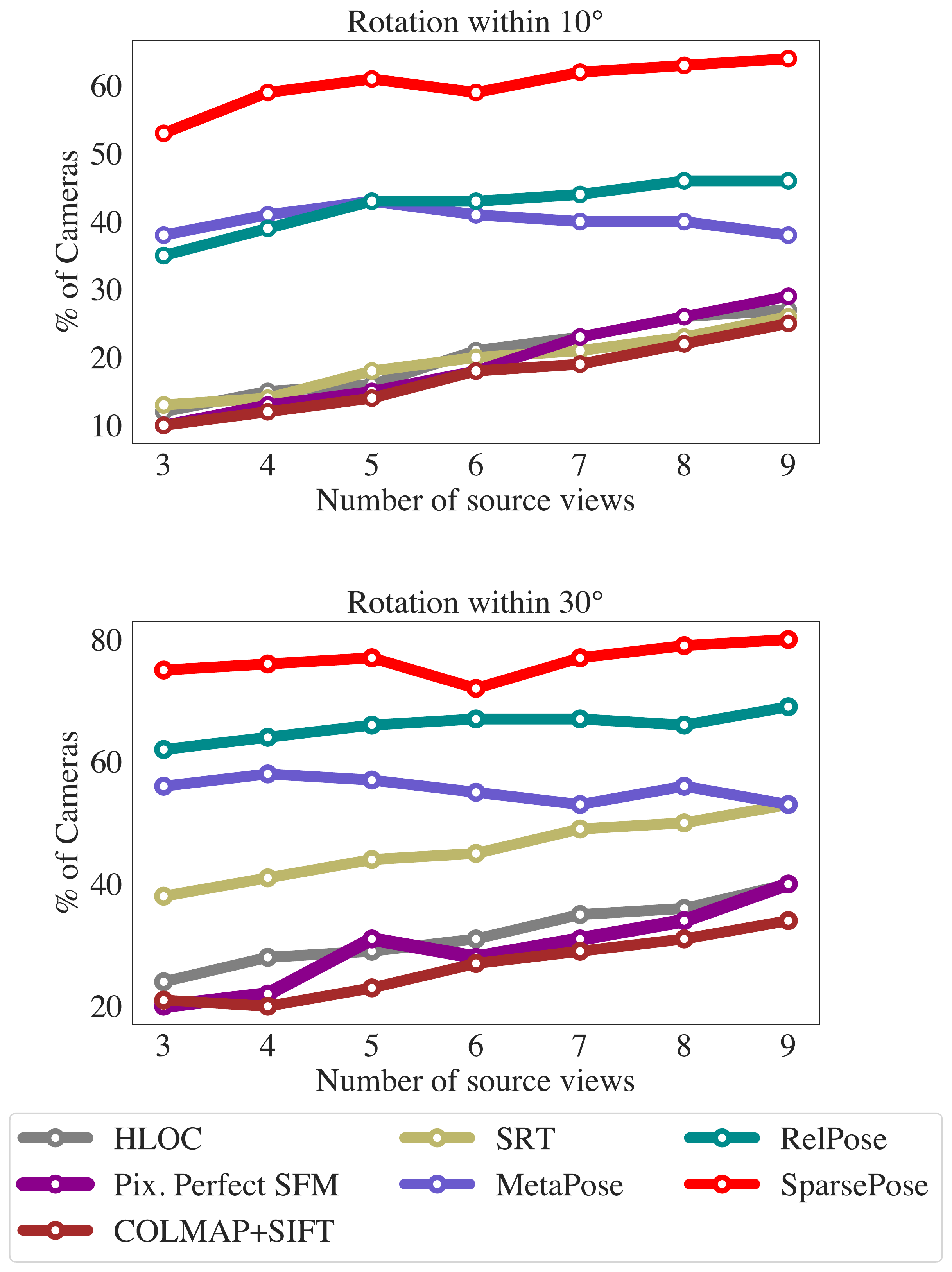}
    \caption{
    \textbf{Evaluating on the percentage
    of cameras accurately predicted with $10^\circ$ and $30^\circ$ 
    thresholds.}
    }
    \label{fig:rot-threshold}
\end{figure}


\section{Baseline details}

\paragraph{MetaPose}
For the MetaPose baseline \cite{metapose}, we used the initialization from SparsePose, and adapt the MetaPose architecture in the officially released code to
perform pose updates on the current updates.
We do not utilize the human-specific information proposed, since our data is more general than the data used to evaluate MetaPose.
We train MetaPose on the same subset of CO3D \cite{co3d} that was used in training \methodNameAbbrev.

\paragraph{Scene Representation Transformer (SRT)}
SRT \cite{srt} proposes to learn a prior over the 3D geometry from data implicitly by learning a ``set-latent scene representation'' from sparse or dense images of the scene using transformer encoder and decoder layers.
Although SRT does not learn a direct 3D geometry of the scene, it does learn a prior over the 3D geometry, as it can perform novel-view synthesis.
To adapt SRT to our evaluation protocol, we add an additional 3-layer MLP that is trained to predict the relative rotations and translations for the input image sequence.
We train the ``unposed'' version of SRT, and add an additional MLP (with 3-hidden layers) to predict rotations and translations, and we train the method with the the same training dataset and loss as \methodNameAbbrev.
For training, we use the same hyperparameters as suggested in the original paper.

\paragraph{Heirarchical Localization (HLOC)}
For HLOC \cite{hloc}, we use the officially released code from 
\url{https://github.com/cvg/Hierarchical-Localization}, which uses SuperPoint 
\cite{superpoint} for generating correspondances and SuperGlue \cite{superglue}
for image matching. 

\paragraph{COLMAP + SIFT}
For the COLMAP baseline with SIFT features, we used the officially released code
from HLOC \url{https://github.com/cvg/Hierarchical-Localization}, which supports  SIFT image features.

\paragraph{RelPose}
For the RelPose baseline, we used the officially released code from 
\url{https://github.com/jasonyzhang/relpose}, and trained on the same dataset 
used to train \methodNameAbbrev. We use the default RelPose hyperparameters.

\paragraph{Pixel Perfect SFM}
For the Pixel Perfect SFM \cite{pixelperfectsfm}, we used the officially 
released codebase from \url{https://github.com/cvg/pixel-perfect-sfm}.
\section{More implementation details}

\begin{table}[h!]
\begin{small}
    \centering
    \begin{tabular}{l c}
        \toprule
        \textbf{Hyperparameter} &  \textbf{Value} \\
        \midrule
        Number of training steps & 500,000 \\
        Number of source views during step & $\mathcal{U}[3,9]$ \\
        Number of sequences sampled per step & 1 \\
        Choice of $\mathcal{E}_{\text{init}}$ & DINO \cite{dino} \\
        Architecture of $\mathcal{E}_{\text{init}}$ & \texttt{ViT-B/8} \cite{vit} \\
        Number of heads $\mathcal{T}_{\text{init}}$ & 8 \\
        Number of heads $\mathcal{T}_{\text{refine}}$ & 2 \\
        Number of hidden dim. $\mathcal{T}_{\text{init}}$, $\mathcal{T}_{\text{refine}}$ & 2048 \\
        Number of hidden layers $\mathcal{N}_{\text{init}}$, $\mathcal{N}_{\text{pose}}$  & 3 \\
        Number of hidden dim. $\mathcal{N}_{\text{init}}$, $\mathcal{N}_{\text{pose}}$  & 512 \\
        Activation for $\mathcal{N}_{\text{init}}$, $\mathcal{T}_{\text{init}}$, $\mathcal{T}_{\text{refine}}$, $\mathcal{N}_{\text{pose}}$ & GELU \\ 
        Number of \textsc{LSTM} steps & 10 \\
        Optimizer & Adam \cite{adam} \\
        Learning rate & $10^{-4}$ \\
        Learning rate decay iterations & 250,000 \\
        Learning rate decay factor & 10\\
        \bottomrule
    \end{tabular}
    \caption{
    \textbf{Hyperparameters and implementation details.} 
    These hyperparameters are shared through all experiments for \methodNameAbbrev, unless stated otherwise.
    }
    \label{tab:hyperparameters}
\end{small}
\end{table}

\section{Qualitative results}

\begin{figure*}
    \centering
    \includegraphics[width=\linewidth]{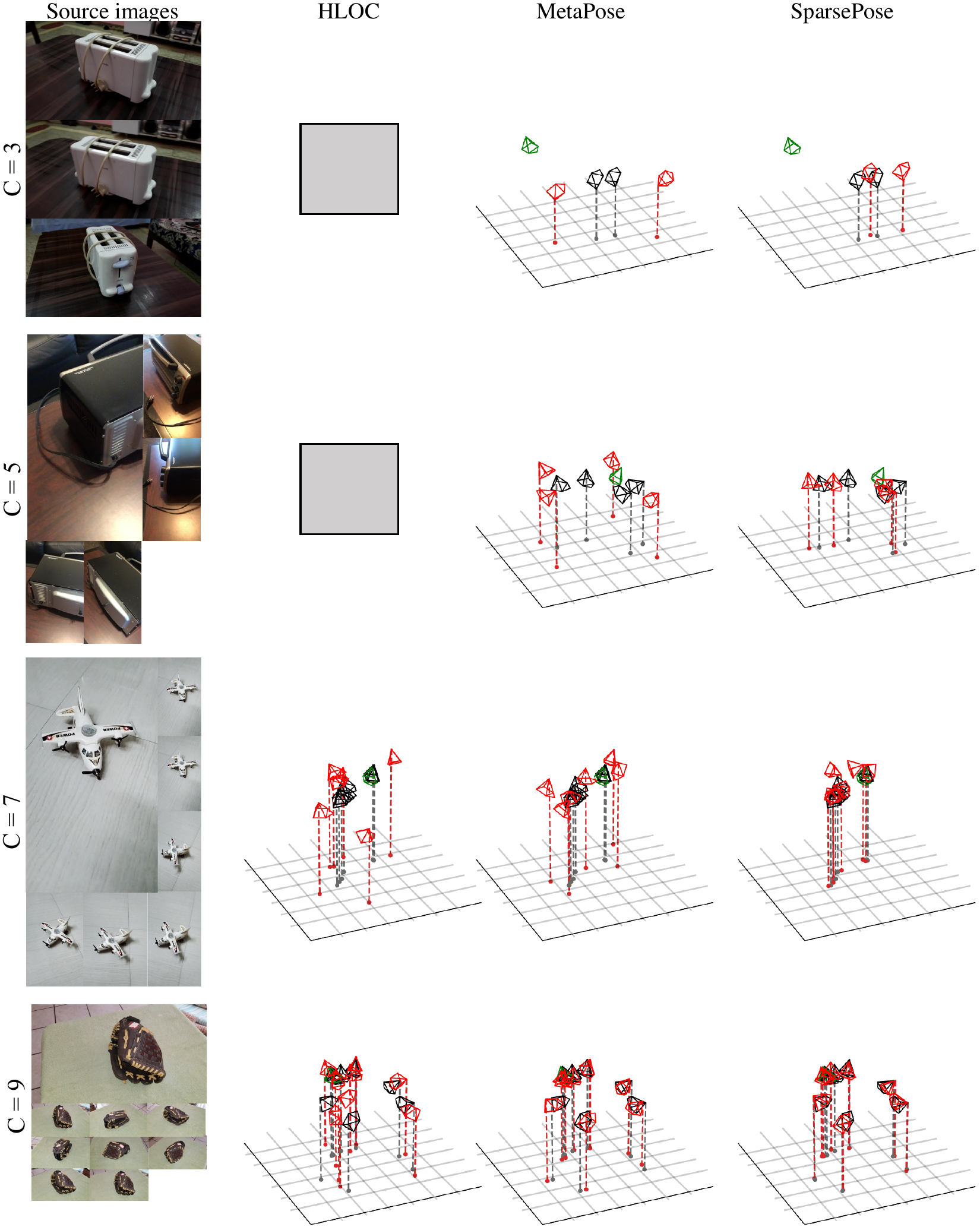}
    \caption{
    \textbf{More qualitative results for the predicted camera poses.}
    The camera centers are projected to the $x-y$ plane for easy visual comparison. The ground truth poses are shown in black, predicted poses in \textcolor{red}{red}, and the first camera for each sequence (used to align predictions) in \textcolor{green}{green}. Gray boxes indicate failure to converge.
    }
    \label{fig:supp_poses}
\end{figure*}

\begin{figure*}
    \centering
    \includegraphics[width=\linewidth]{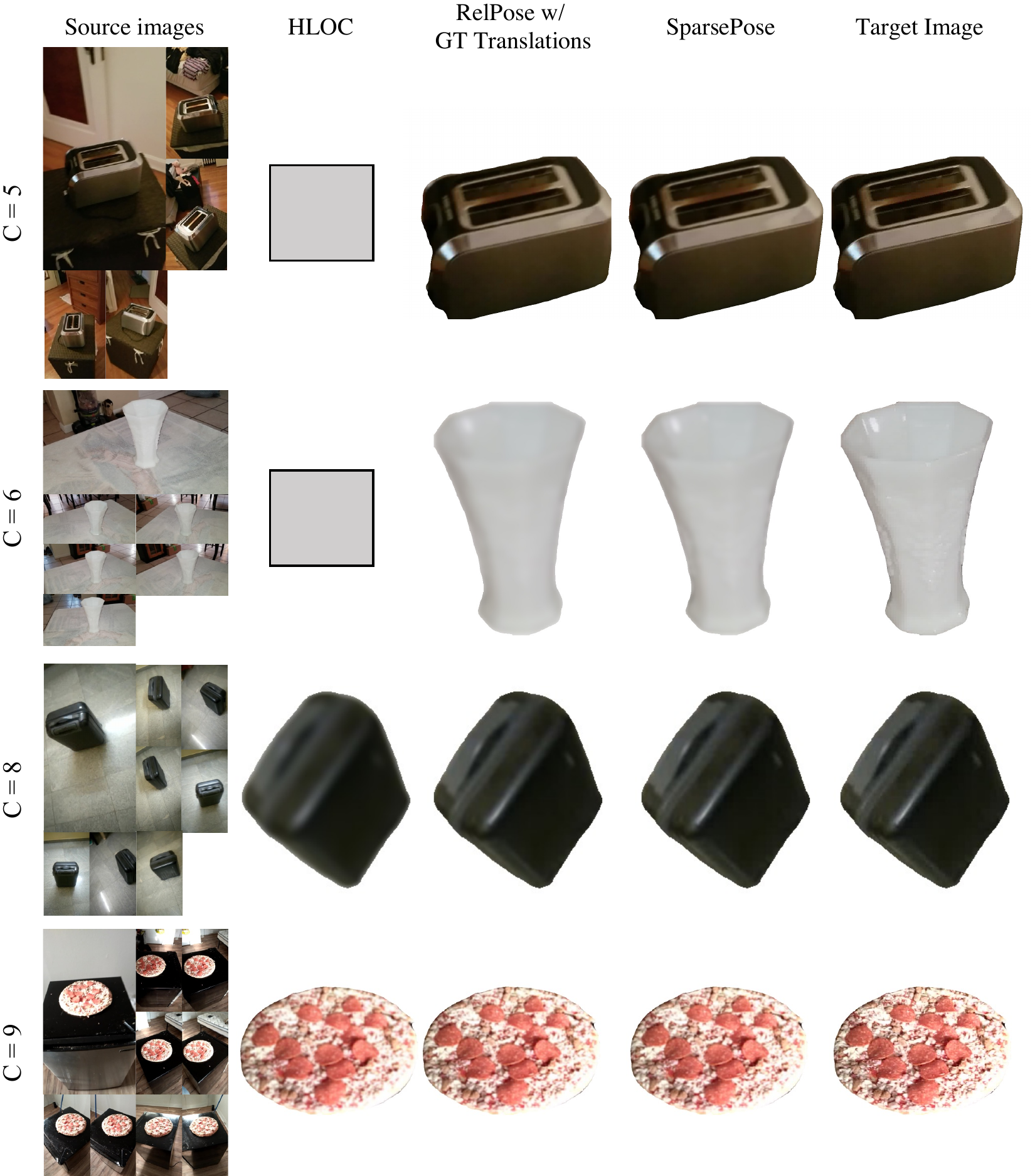}
    \caption{\textbf{More qualitative renders from a sparse set of unposed images.}}
    \label{fig:supp_render_1}
\end{figure*}

\begin{figure*}
    \centering
    \includegraphics[width=\linewidth]{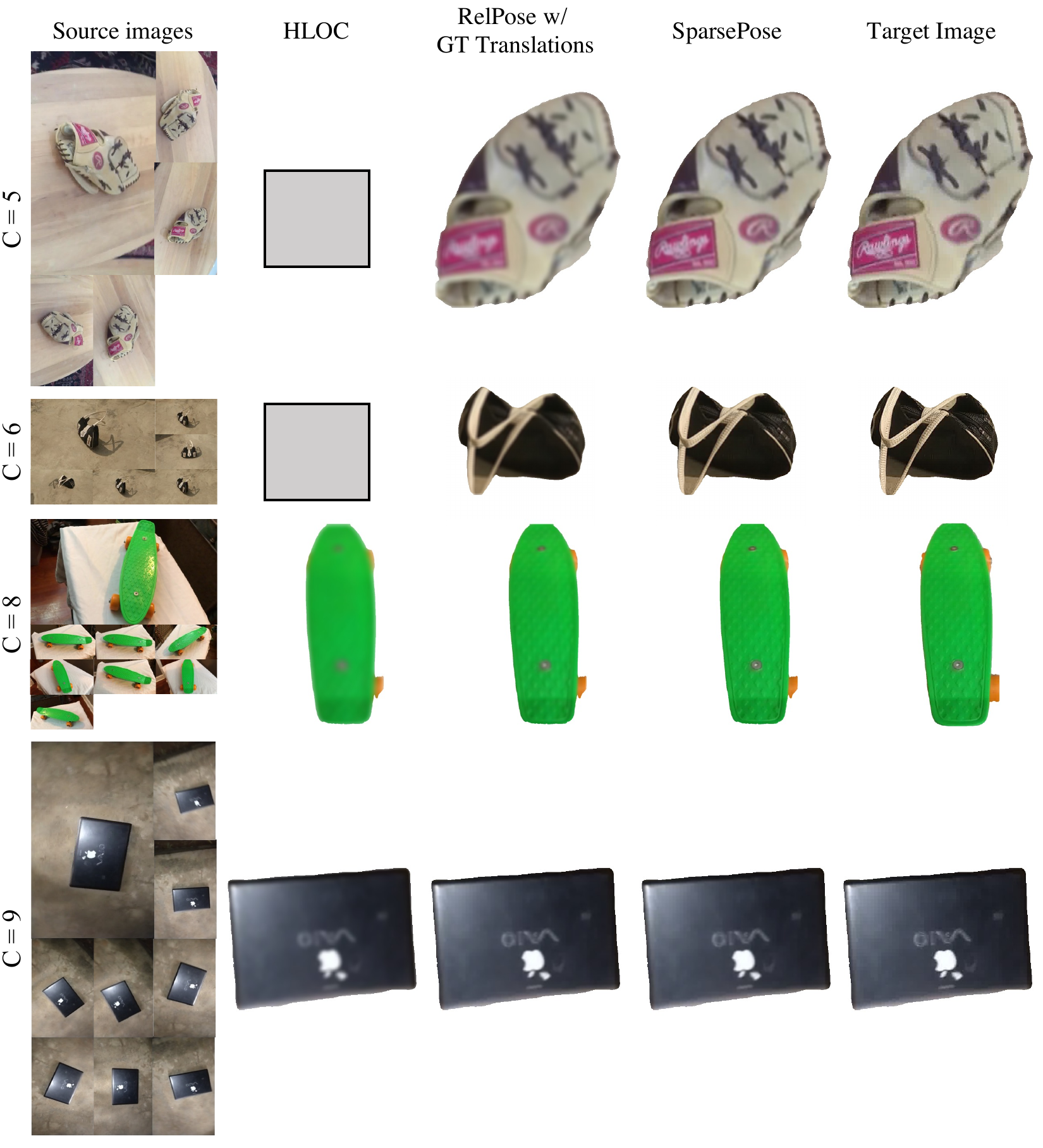}
    \caption{\textbf{More qualitative renders from a sparse set of unposed images.}}
    \label{fig:supp_render_2}
\end{figure*}

In Figure~\ref{fig:supp_poses} we provide additional qualitative results with different numbers of source views and visualize the predicted camera poses by our method compared to baselines. 
We also include additional qualitative novel-view synthesis results
for different categories over different numbers of source views in Figure~\ref{fig:supp_render_1} and Figure~\ref{fig:supp_render_2}. 
In both cases, we see that \methodNameAbbrev predicts more accurate
camera poses, resulting in higher quality novel-view synthesis compared to other baselines. 

\end{document}